\documentclass{article} % For LaTeX2e
\usepackage{iclr2027_conference,times}

\usepackage{amsmath,amsfonts,bm}

\def\eqref#1{equation~\ref{#1}}
\def\1{\bm{1}}

\DeclareMathAlphabet{\mathsfit}{\encodingdefault}{\sfdefault}{m}{sl}
\SetMathAlphabet{\mathsfit}{bold}{\encodingdefault}{\sfdefault}{bx}{n}

\usepackage{kotex}

\usepackage{amsmath,amssymb,amsfonts,bm}
\usepackage{graphicx}
\usepackage{booktabs}
\usepackage{algorithm}
\usepackage{algorithmic}
\usepackage{hyperref}
\usepackage{url}
\usepackage{xspace}
\usepackage[most]{tcolorbox}
\usepackage{tikz}
\usetikzlibrary{calc}
\usepackage{fontawesome}
\usepackage{enumitem}
\usepackage{makecell}
\usepackage{booktabs}
\usepackage{multirow}
\usepackage[table]{xcolor}
\usepackage{wrapfig}
\usepackage{graphicx}
\usepackage{caption}
\usepackage{mdframed}

\definecolor{oursblue}{RGB}{224, 238, 250}

\newcommand{\name}{PLRM\xspace}

\newcommand{\ie}{\emph{i.e.,}\xspace}

\definecolor{insightteal}{RGB}{34, 139, 139}   % A sophisticated, medium-dark teal
\definecolor{insightback}{RGB}{240, 248, 248}   % A very light, complementary cyan-white

\definecolor{paleblue}{RGB}{70,130,180}
\definecolor{linkcolor}{RGB}{70,130,180}
\definecolor{plugincolor}{RGB}{220,40,40}

\hypersetup{
    colorlinks=true,
    citecolor=paleblue,   % Changes the color of citation links
    linkcolor=linkcolor,   % Changes the color of normal internal links (optional)
    urlcolor=magenta  % Changes the color of URLs (optional)
}

\newcommand{\lightbulbicon}{%
  \begin{tikzpicture}[baseline=-0.5ex]
    \draw[fill=white, draw=insightteal, thick] (0,0) circle (1.5ex);
    \node[scale=0.8, color=insightteal] at (0,0) {\faLightbulbO~};
  \end{tikzpicture}%
}

\newtcolorbox{customblockquote}{
  colframe=insightteal,
  colback=insightback,
  boxrule=0pt,
  left=5pt,
  right=4pt,
  top=5pt,
  bottom=3pt,
  arc=0pt,
  breakable,
  before skip=1.2\baselineskip,
  after skip=0.7\baselineskip,
  left skip=0pt,
  right skip=0pt,
  enhanced jigsaw,
  frame hidden,
  overlay={
    \draw[insightteal, line width=2pt] 
      ([yshift=1pt]frame.north west) -- (frame.south west);
    \node[inner sep=0pt] at ([xshift=0pt, yshift=-1.3pt]frame.north west) {\lightbulbicon};
  },
  fontupper=\fontfamily{lmr}\selectfont,
  boxsep=1pt,
}

\definecolor{greydark}{RGB}{90, 90, 90} 
\definecolor{greylight}{HTML}{F1F3F8}

\newtcolorbox{greycustomblock}{
  colframe=greydark,        % Left line color
  colback=greylight,        % Background color
  boxrule=1pt,              % Line width
  left=2.5pt,               % Inner left margin
  right=3pt,                % Inner right margin
  top=5pt,                  % Inner top margin
  bottom=3pt,               % Inner bottom margin
  arc=0pt,                  % No rounded corners
  breakable,                % Allow breaking across pages
  before skip=0.2\baselineskip, % Vertical space before
  after skip=0.2\baselineskip,  % Vertical space after
  left skip=0pt,            % Left skip (adjust as needed)
  right skip=0pt,           % Right skip (adjust as needed)
  enhanced jigsaw,          % Enhanced jigsaw for precise control
  frame hidden,             % Hide the default frame
  overlay={                 % Custom overlay for left line
    \draw[greydark, line width=2pt]
      ([yshift=-1pt]frame.north west) -- ([yshift=1pt]frame.south west); % Adjusting the y coordinates to match exactly
  },
  fontupper=\selectfont, % Font size and style for the table
}

\mdfdefinestyle{mathblockframe}{
  linecolor=gray!40,
  linewidth=0.5pt,
  backgroundcolor=gray!8,
  roundcorner=3pt,
  innertopmargin=5pt,
  innerbottommargin=5pt,
  innerleftmargin=3pt,
  innerrightmargin=3pt,
  skipabove=3pt,
  skipbelow=3pt,
}

\newmdtheoremenv[style=mathblockframe]{theorem}{Theorem}[section]
\newmdtheoremenv[style=mathblockframe]{proposition}{Proposition}
\newmdtheoremenv[style=mathblockframe]{corollary}[theorem]{Corollary}
\newmdtheoremenv[style=mathblockframe]{observation}[theorem]{Observation}

\title{%
\protect\parbox{\textwidth}{%
\centering
Parallelizable Probabilistic Latent Reasoning
for Diverse Reasoning Trajectories
\par
\vspace{14pt}
}}

\author{%
\makebox[\textwidth][c]{%
\begin{tabular}{@{}c@{\hspace{18pt}}c@{\hspace{18pt}}c@{}}
\textbf{Seokhyun Lee} &
\textbf{Seongjun Lee} &
\textbf{Changhee Lee}\thanks{Corresponding author.}
\\[5pt]
\normalfont Korea University &
\normalfont Korea University &
\normalfont Korea University
\\[3pt]
{\normalfont\footnotesize\texttt{nshuhsn@korea.ac.kr}} &
{\normalfont\footnotesize\texttt{pyoung7307@korea.ac.kr}} &
{\normalfont\footnotesize\texttt{changheelee@korea.ac.kr}}
\end{tabular}%
}%
}

\iclrfinalcopy

\begin{document}

\maketitle
\lhead{Preprint}

\begin{abstract}
Language models relying on verbalized Chain-of-Thought (CoT) are constrained to token-by-token generation in the language space, where many tokens serve textual coherence rather than core reasoning. This entangles genuine differences in logical trajectories with superficial variations in verbalization. Moving reasoning into a compact, continuous latent space offers a promising alternative for efficiently representing and exploring diverse reasoning trajectories. However, current latent models remain largely deterministic, capturing only a single reasoning trajectory rather than modeling a distribution over multiple valid alternatives. While stochastic approaches parameterize distributions over latent reasoning steps, they frequently suffer from unstable optimization and limited empirical performance. Furthermore, their reliance on autoregressive latent prediction during training precludes effective parallelization, leading to prohibitively long training times.
To overcome these bottlenecks, we propose \textit{Probabilistic Latent Reasoning Models} (\name), a parallelizable variational framework for probabilistic latent reasoning. \name adopts a JEPA-style learning framework in which the posterior infers latent distributions in parallel from observed reasoning trajectories, while the prior learns to predict the subsequent latent distribution via teacher forcing. 
To model heterogeneous reasoning processes, we introduce a discrete strategy variable together with a mixture-based prior, allowing the model to preserve the Gaussian-mixture structure induced by multiple distinct reasoning trajectories rather than collapsing them into a single Gaussian. Across diverse benchmarks, \name achieves performance competitive with deterministic latent reasoning models on single-trajectory datasets while enabling efficient training and consistently outperforming competing stochastic baselines on multi-trajectory datasets. Codes are available \href{https://github.com/nshuhsn/PLRM}{\underline{here}}.
\end{abstract}

\section{Introduction}\label{sec:intro}
Autoregressive language models have developed strong reasoning capabilities by learning to generate chain-of-thought (CoT) traces in natural language~\citep{wei_COT_NeurIPS2022,chung_CoTSFT_JMLR2024}.
This success has established language as the dominant interface for both supervising and executing intermediate reasoning. A key advantage is that next-token prediction exposes reasoning as an explicit generative trajectory, enabling the exploration of multiple hypotheses and solution paths for tasks such as mathematical proof search and scientific hypothesis generation~\citep{wang_Selfconsistency_ICLR2023,yao_ToT_NeurIPS2023,trinh_AlphaGeometry_Nature2024,gottweis_CoScientist_Nature2026}. It also enables reinforcement learning by optimizing generated reasoning trajectories from reward feedback~\citep{ouyang_RLHF_NeurIPS2022}.

However, despite its favorable training properties and strong empirical performance, language is not necessarily an efficient substrate for reasoning. Many tokens primarily serve linguistic coherence rather than core reasoning, yet current Large Language Models (LLMs) devote comparable computation to both reasoning-critical and purely linguistic tokens~\citep{hao_coconut_COLM2025}. This is not only introduces substantial computational overhead, but also entangles genuine differences in reasoning logic with superficial variations in how the same reasoning process is verbalized~\citep{kuhn_SemantiUncertainty_ICLR2023}. Moreover, each reasoning step in language space is restricted to a discrete choice from a vocabulary of size $v$, limiting each sampled token to at most $\log_2 (v)$ bits of information~\citep{gozeten_CoT2_ICLR2026}.

% whereas continuous high-dimensional representations can express richer information

These limitations raise a fundamental question:~\emph{in what representational space should reasoning unfold?}
Latent reasoning models address this question by replacing language tokens with continuous representations.
Coconut~\citep{hao_coconut_COLM2025} uses curriculum learning to progressively replace CoT tokens with continuous latent representations, while CODI~\citep{shen_Codi_EMNLP2025} and Sim-CoT~\citep{wei_SIMCoT_ICLR2026} employ self-distillation and language reconstruction to stabilize the transition to latent reasoning. These approaches achieve competitive performance with CoT supervised fine-tuning~(CoT-SFT) methods while requiring substantially fewer reasoning steps. However, this transition from language to latent space sacrifices several desirable properties of SFT. Most notably, existing latent reasoning models largely rely on deterministic latent transitions, reducing diverse reasoning paths to a single trajectory and thereby limiting their ability to model reasoning as a distribution over trajectories.

In this work, we introduce \name, a variational framework for probabilistic latent reasoning that moves reasoning from language to continuous latent space while preserving the key properties that make CoT-SFT effective.
Rather than merely replacing tokens with latent representations, \name adopts a JEPA-style predictive objective: a posterior constructs target latent distributions from verbal CoT traces, while a problem-conditioned prior learns to predict them.
By jointly optimizing both end-to-end, \name encourages the latent representations to retain core reasoning information while discarding incidental variation in verbalization.
We further show that multiple valid reasoning trajectories for the same problem can induce a multimodal target distribution, and introduce a mixture-based prior to model this multimodality without collapsing it into a single Gaussian.

Experimentally, \name performs strongly across mathematical and commonsense reasoning tasks, substantially outperforming existing stochastic latent reasoning models while remaining competitive with deterministic latent reasoning methods under more efficient training.
Performance is particularly strong on GSM8K-Aug-NL~\citep{deng_GSM8K_arxiv2023} and CommonsenseQA~\citep{talmor_commonsenseQA_NAACL2019}, where many CoT tokens serve verbalization rather than core reasoning, consistently capturing core reasoning information in latent representations. To evaluate distributional modeling under diverse reasoning trajectories, we further construct the HiddenBranch dataset from ProofWriter~\citep{tafjord_proofwriter_ACL2021}, where each problem admits multiple reasoning paths and answers. On this dataset, \name substantially outperforms CoT-SFT and all stochastic latent reasoning baselines, demonstrating its ability to effectively model distributions over diverse reasoning trajectories.

% \begin{customblockquote}
% \define{Contribution} (i) We formulate probabilistic latent reasoning within a variational formulation that supports distributional modeling over reasoning trajectories, stepwise supervision, and parallel training.
% (ii) This formulation leads to a JEPA-style learning framework that jointly optimizes posterior target distributions and a problem-conditioned prior, encouraging latent representations to capture core reasoning information. (iii) We show that multiple valid reasoning trajectories can induce multimodal target distributions and introduce a mixture-based prior to preserve this structure rather than collapsing it into a single Gaussian. (iv) We show empirically that \name achieves competitive performance on single-trajectory datasets with efficient training, and outperforms SFT and stochastic reasoning baselines on multi-trajectory datasets
% \end{customblockquote}

\section{Desiderata for Latent Reasoning} \label{sec:desiderata}
In this section, we first identify the desirable properties of conventional supervised fine-tuning, then examine how latent reasoning changes this paradigm, highlighting both the benefits it introduces and the properties it may sacrifice.

\subsection{Properties from  Conventional Supervised Fine-Tuning} 
Consider a language model that receives a problem $x$, generates a verbalized reasoning trace~(CoT) $\boldsymbol{\tau} = (\tau_1,\cdots,\tau_T)$, and then predicts the final answer $y$. Conventional supervised fine-tuning can be viewed as learning the following joint distribution:
\begin{equation} \label{eq:sft}
p(y,\boldsymbol{\tau}\mid x) = p(y\mid \boldsymbol{\tau},x)\cdot p(\boldsymbol{\tau}\mid x)    
\end{equation}
This formulation provides three desirable properties. \textbf{First}, next-token prediction is inherently stochastic, allowing the model to represent uncertainty and explore multiple reasoning trajectories through sampling. This breadth-wise exploration is essential for discovering diverse solutions ~\citep{wang_Selfconsistency_ICLR2023} and also underlies the basis for reinforcement-learning-based optimization, where trajectories are sampled and optimized according to their rewards~\citep{ouyang_RLHF_NeurIPS2022}.
\textbf{Second}, every observed reasoning token serves as a prediction target, providing dense, fine-grained supervision throughout the reasoning trajectory~\citep{bengio_PLM_JMLR2003}. \textbf{Finally}, teacher forcing on the ground-truth trajectory enables parallel training, a property that has been fundamental to the efficient scaling of modern language models~\citep{vaswani_Transformer_NeurIPS2017}. 

\subsection{Shifting Reasoning from Language Space to Latent Space}
To improve the inefficiency of language-space reasoning, latent reasoning models replace verbalized reasoning steps with a sequence of continuous latent variables~\citep{hao_coconut_COLM2025}.
Concretely, conventional language-space reasoning computes a vocabulary distribution from the final-layer hidden representation $h_t$ and samples the next reasoning token: $h_t = \mathrm{Transformer}(x,\tau_{<t}),
\qquad
p(\tau_t\mid x,\tau_{<t})
=
\mathrm{Softmax}(W_{\mathrm{out}}h_t),$ where $W_{\mathrm{out}}$ denotes the output head of a language model.
Latent reasoning instead maps the hidden representation directly to a continuous latent variable through a learned projection, $z_k=g_\psi(h_k)$, and uses the resulting latent variable as an intermediate reasoning step. Repeating this process provides a latent reasoning trajectory $\mathbf z=(z_1,\ldots,z_K)$ that replaces the verbalized reasoning trace $\boldsymbol{\tau}$, leading to the generic factorization $p(y,\mathbf z\mid x)=p(y\mid\mathbf z,x)p(\mathbf z\mid x)$.

\textbf{Deterministic Latent Reasoning and Its Pitfalls.}
While latent reasoning removes intermediate verbalization, the three desirable properties above are no longer preserved by construction. A deterministic projection produces a single latent trajectory, foregoing probabilistic modeling of alternative reasoning paths and limiting the latent space's ability to represent diverse underlying reasoning processes. 
Moreover, the absence of target latent variables prevents dense supervision and teacher forcing, heavily bottlenecking the training process with sequential rollouts~\citep{hao_coconut_COLM2025,shen_Codi_EMNLP2025}. SIM-CoT~\citep{wei_SIMCoT_ICLR2026} partially addresses the supervision gap by reconstructing $\boldsymbol{\tau}$ from each latent step, but the underlying latent trajectories remain deterministic and still require slow, sequential execution during training.

% \textbf{Potential and Challenges of Stochastic Latent Reasoning.}
% Stochastic latent reasoning methods instead predicts $(\mu_k,\sigma_k)=g_\psi(h_k)$ and samples $z_k\sim\mathcal N(\mu_k,\sigma_k^2)$ to model uncertainty over alternative trajectories. Existing methods typically guide these distributions using heuristically constructed or externally derived targets: CoLaR~\citep{tan_CoLaR_NeurIPS2026} uses scaled sums of CoT token embeddings, while ReGuLaR~\citep{wang_regular_arXiv2026} derives Gaussian priors from rendered CoT representations. Since its posterior generates the latent trajectory sequentially, training cannot be parallelized across latent steps. More fundamentally, such targets are not explicitly optimized to preserve answer-relevant core reasoning information or to remain predictable from $x$ alone. Consequently, stochasticity alone does not ensure that the learned latent distribution captures meaningful variation in reasoning.

\textbf{Potential and Challenges of Stochastic Latent Reasoning.}
Unlike deterministic latent reasoning, stochastic modeling can better exploit the latent space to represent a distribution over alternative reasoning trajectories, distinguishing genuine differences in reasoning logic from superficial variations in verbalization. This enables breadth-wise exploration directly in the core reasoning space and provides a stochastic foundation for reinforcement learning over latent reasoning trajectories. A straightforward approach is to predict Gaussian parameters from the hidden representation and sample each latent variable as
\begin{equation}
(\mu_k,\sigma_k)=g_\psi(h_k),
\qquad
z_k\sim\mathcal N(\mu_k,\sigma_k^2).
\end{equation}
However, introducing stochasticity alone does not determine what a meaningful latent reasoning distribution should represent. A useful target distribution should retain answer-relevant reasoning information while excluding incidental information that cannot be inferred from $x$. Existing methods typically impose guidance through heuristically constructed or externally derived targets: CoLaR~\citep{tan_CoLaR_NeurIPS2026} uses scaled sums of CoT token embeddings, while ReGuLaR~\citep{wang_regular_arXiv2026} derives Gaussian priors from rendered CoT representations using a frozen visual encoder and a trainable adapter. Such targets are not explicitly learned to preserve answer-relevant core reasoning information while remaining predictable from $x$, and some approaches still require sequential latent rollout.
Thus, the key challenge is not merely to make latent reasoning stochastic, but to learn a meaningful and predictable distribution over reasoning trajectories.

\section{Probabilistic Latent Reasoning Models} \label{sec:problem}

% In this section, we first revisit how language models reason with Chain-of-Thought (CoT) and identify the key properties that make this paradigm effective. We then use these properties to motivate the desiderata for latent reasoning and introduce our new framework, the Probabilistic Latent Reasoning Models (\name). 

Motivated by above, we introduce Probabilistic Latent Reasoning Models (\name), a variational framework that jointly learns target latent distributions and a problem-conditioned prior to predict them. We first develop a variational formulation that preserves the three desirable properties of conventional supervised fine-tuning. We then show that this formulation naturally takes the form of a JEPA-style learning framework, in which target latent distributions capture core reasoning information while abstracting away incidental variations in verbalization. Finally, we extend the framework to settings with multiple heterogeneous reasoning strategies by introducing an explicit strategy variable and a mixture-based prior capable of representing distinct modes of the reasoning distribution.

\subsection{Variational Latent Reasoning} 
% Guided by these desiderata, we seek a latent reasoning formulation that removes the inefficiency of reasoning in the language space while preserving its key advantages. We begin by introducing a latent reasoning trajectory, $\mathbf z=(z_1,\ldots,z_K)$, that represents the underlying reasoning process. The verbalized reasoning trace $\tau$ and the final answer $y$ are then modeled as being generated from this latent trajectory, yielding

We define $\mathbf z$ as the latent reasoning trajectory underlying both the verbalized trace $\boldsymbol{\tau}$ and the final answer $y$, following the generative structure $x \rightarrow \mathbf z \rightarrow \{\boldsymbol{\tau},y\}$. Under this formulation, the trace $\boldsymbol{\tau}$ is treated as one possible verbal realization from $\mathbf z$, \ie $\boldsymbol{\tau}\sim p_\theta(\boldsymbol{\tau}\mid\mathbf z)$. Accordingly, the joint distribution in Eq.~\ref{eq:sft} can be reformulated by marginalizing over the latent reasoning trajectory as
\begin{equation}
p_\theta(y,\boldsymbol{\tau} \mid x)
=
\int
p_\theta(\boldsymbol{\tau}\mid \mathbf z)\,
p_\theta(y\mid \mathbf z,x)\,
p_\theta(\mathbf z\mid x)\,dz .
\end{equation}
Here, $p_\theta(\mathbf z\mid x)=\prod_{k=1}^{K}p_\theta(z_k\mid x,z_{<k})$ represents the distribution over possible latent reasoning processes for a given problem. We retain $x$ in $p_\theta(y\mid\mathbf z,x)$ for computational efficiency: since answer generation follows the existing $(x,\mathbf z)$ context, the model can continue decoding directly from the current KV cache without recomputing the context representations.

% We omit $x$ from $p_\theta(\boldsymbol{\tau}\mid\mathbf z)$ to ensure that the verbalized trace is reconstructed from the latent trajectory itself. In contrast, 

To benefit from fine-grained supervision and parallel training in the latent space, we introduce a factorized variational posterior $q_\phi(\mathbf z\mid x,\boldsymbol{\tau})
=\prod_{k=1}^{K}q_\phi(z_k\mid x,\boldsymbol{\tau})$. Because each posterior distribution is inferred directly from the observed reasoning trace, every latent step is associated with the target distribution without requiring sequential posterior inference. In our basic formulation, we parameterize both the posterior and prior at each latent step as Gaussian distributions:
\begin{equation}
q_\phi(z_k\mid x,\boldsymbol{\tau})
=
\mathcal{N}\!\left(
\mu_k^q,
(\sigma_k^q)^2
\right),
\qquad
p_\theta(z_k\mid x,z_{<k})
=
\mathcal{N}\!\left(
\mu_k^p,
(\sigma_k^p)^2
\right).
\end{equation}
The framework is then jointly trained by optimizing the following Evidence Lower Bound~(ELBO) of $\log p_\theta(y, \boldsymbol{\tau} \mid x)$:
\begin{equation}\label{eq:elbo}
% \log p_\theta(y, \boldsymbol{\tau} \mid x) \ge\;&
\underbrace{
\mathbb{E}_{q_\phi(\mathbf z\mid x,\boldsymbol{\tau})}
\left[\log p_\theta(\boldsymbol{\tau}\mid \mathbf z)\right]
}_{\text{CoT reconstruction}}
+
\underbrace{
\mathbb{E}_{q_\phi(\mathbf z\mid x,\boldsymbol{\tau})}
\left[\log p_\theta(y\mid \mathbf z,x)\right]
}_{\text{answer prediction}}
% \nonumber\\
-
\underbrace{
D_{\mathrm{KL}}
\left(
q_\phi(\mathbf z\mid x,\boldsymbol{\tau})
\,\|\, 
p_\theta(\mathbf z\mid x)
\right)
}_{\text{distribution alignment}} .
\end{equation}
\begin{customblockquote}
\textbf{Takeaway.} Importantly, this variational formulation moves reasoning from language space to latent space while maintaining three desired properties of conventional supervised fine-tuning. \textbf{First}, both the posterior $q_\phi(\mathbf z\mid x,\boldsymbol{\tau})$ and the prior $p_\theta(\mathbf z\mid x)$ model distributions over latent trajectories, preserving stochasticity over alternative reasoning paths. \textbf{Second}, the posterior provides dense supervision by inferring a target distribution for every latent step directly from the observed verbal reasoning trace. \textbf{Finally}, since the KL objective is evaluated on posterior-generated latent trajectories, the prior performs teacher-forced next-latent distribution prediction, allowing all latent steps to be trained in parallel.
\end{customblockquote}

Having moved reasoning from language space to a continuous latent space, a natural question arises: \emph{what makes a useful latent reasoning representation?} 
We argue that such a representation should satisfy two basic requirements.
First, because reasoning ultimately serves to solve the problem, the latent trajectory should preserve sufficient information to predict the correct answer.
Second, because the latent trajectory must be generated from the problem alone at inference time, the latent trajectory should remain predictable from the problem without access to the verbalized reasoning trace.
Together, these requirements favor latent representations that capture core, problem-relevant reasoning information while abstracting away incidental details arising from a particular verbalization.

\subsection{Joint Distribution Predictive Architecture}
Our formulation in Eq.~\ref{eq:elbo} captures these requirements through three complementary objectives:
\begin{enumerate}[label=\textcircled{\scriptsize\arabic*},leftmargin=1.7em,topsep=0pt, partopsep=0pt,itemsep=0pt, nosep]
    \item $p_\theta(\boldsymbol{\tau}\mid\mathbf{z})$: preserve information about the verbalized reasoning trace $\boldsymbol{\tau}$, \vspace{1mm}
    \item $p_\theta(y\mid\mathbf{z},x)$: support prediction of the final answer $y$, and \vspace{1mm}
    \item $D_{\mathrm{KL}}(q_\phi(\mathbf z\mid x,\boldsymbol{\tau})
\!\,\|\, 
p_\theta(\mathbf z\mid x))$: remain predictable from the prior conditioned on the problem $x$.
\end{enumerate}

While the first two objectives ground the latent representations in task-relevant semantics, the third objective shapes the structure of the latent space itself.
Specifically, the posterior acts as a target encoder that provides latent target distributions, while the prior acts as a predictor that learns to match these targets without access to the verbalized reasoning trace. This naturally gives rise to a JEPA-style framework,~\citep{baevski_Data2vec_ICML2022, assran_JEPA_CVPR2023}, with the distinction that both the target and predicted representations are probability distributions rather than deterministic embeddings. 

In the spirit of JEPA, which emphasizes predictive representations over reconstructing every detail of the input, our framework encourages the latent representation to capture core reasoning information while abstracting away incidental details of its verbal realization.
The following proposition provides an information-theoretic interpretation of the role of the problem-conditioned prior $p_\theta(\mathbf z\mid x)$:

\begin{proposition}[Problem-Conditioned Information Bottleneck]~\label{pp:IB} Considering the expectation of  distribution alignment over the data distribution $p(x,\boldsymbol{\tau})$, we obtain
\begin{align} \label{eq:ib}
\mathbb E_{x,\boldsymbol{\tau}} \big[
D_{\mathrm{KL}}\!\left(
q_\phi(\mathbf z\mid x,\boldsymbol{\tau})
\,\|\,p_\theta(\mathbf z\mid x)
\right) \big]
&=
\mathbb E_{x,\boldsymbol{\tau},\mathbf z}
\left[
\log
\frac{q_\phi(\mathbf z\mid x,\boldsymbol{\tau})}
{\bar q_\phi(\mathbf z\mid x)}
\right]
+
\mathbb E_{x,\mathbf z}
\left[
\log
\frac{\bar q_\phi(\mathbf z\mid x)}
{p_\theta(\mathbf z\mid x)}
\right]
\nonumber\\
&=
I_q(\mathbf z;\boldsymbol{\tau}\mid x)
+
\mathbb E_x\big[
D_{\mathrm{KL}}\!\left(
\bar q_\phi(\mathbf z\mid x)
\,\|\,p_\theta(\mathbf z\mid x)
\right)\big],
\end{align}
where $\bar q_\phi(\mathbf z\mid x)
=
\mathbb E_{\boldsymbol{\tau}\sim p(\boldsymbol{\tau}\mid x)}
[q_\phi(\mathbf z\mid x,\boldsymbol{\tau})]$ denotes the aggregated posterior over latent reasoning trajectories for a given problem $x$.
Thus, as the prior approaches the aggregated posterior, the alignment objective increasingly penalizes information about the particular verbalized trace that is not predictable from the problem. 
\end{proposition} \vspace{-2mm}

Together with answer prediction loss $p_\theta(y\mid\mathbf{z},x)$, this biases $\mathbf z$ toward retaining task-relevant reasoning information while suppressing trace-specific variation.

\textbf{Preventing Representation Collapse.}
To prevent collapse of the target representations, recent JEPA methods employ representation regularization that shapes the embedding distribution toward a standard normal Gaussian~\citep{balestriero_Lejepa_arXiv2025, maes_leworldmodel_ICML2025}. Following this principle, we apply VISReg~\citep{wu_visreg_arXiv2026} to the posterior means $\mu^q$ to stabilize joint posterior--prior training. We regularize $\mu^q$ rather than sampled latent $z$, since Gaussian sampling noise could satisfy the regularization even when the posterior means collapse.

% Overall, the formulation above transfers the desirable properties of conventional CoT supervision into a latent space. So far, however, we have assumed that each prior conditional is represented by a single Gaussian. We next consider the more general setting where the same problem admits multiple heterogeneous reasoning processes, and show that a single-Gaussian prior can be insufficient to match the resulting aggregated posterior targets.

% \textbf{Modeling Heterogeneous Reasoning Strategies.} 
\subsection{Modeling Heterogeneous Reasoning Strategies} \label{subsec:Multimodal} 
As Proposition~\ref{pp:IB} shows, the prior is optimized to match the aggregated posterior $\bar q_\phi(\mathbf z\mid x)$. When a problem $x$ has a single reasoning trace $\boldsymbol{\tau}$,  $\bar q_\phi(\mathbf z\mid x)$ reduces to $q_\phi(\mathbf z\mid x,\boldsymbol{\tau})$, recovering the original distribution-alignment term.
When $x$ admits multiple reasoning trajectories with corresponding verbalized traces $\{\boldsymbol{\tau}_i\}_{i=1}^{N_x}$, marginalizing over all valid traces induces a multimodal target distribution. 
This multimodality can already arise at the first latent step: $\bar q_\phi(z_1\mid x)
=
\mathbb E_{\boldsymbol{\tau}\sim p(\boldsymbol{\tau}\mid x)}
[
q_\phi(z_1\mid x,\boldsymbol{\tau})
] 
=
\sum_{i=1}^{N_x}
p(\boldsymbol{\tau}_i\mid x)\,
\mathcal N(
\mu_{i,1}^q,
(\sigma_{i,1}^q)^2
),$ whereas the basic prior is restricted to a single Gaussian, \ie $p_\theta(z_1\mid x)
=
\mathcal N(
\mu_1^p,
(\sigma_1^p)^2
)$. 

% Because the posterior and prior are jointly optimized under the KL objective, restricting the prior to a single Gaussian leads to two complementary failure modes. 
This mismatch creates two complementary failure modes. 
First, the posterior can reduce the KL divergence by pulling trajectory-specific Gaussian modes toward a common region, collapsing distinct reasoning strategies in the latent space. 
Second, if the posterior preserves these distinct modes, the prior must cover them with a single Gaussian. Under the forward KL, this encourages the prior to assign probability mass across all modes, potentially inflating its variance and placing substantial mass between them.
Thus, a unimodal prior can either \textit{collapse the multimodal structure of the target representation} or \textit{blur it in the predictive distribution}.

\textbf{Strategy-Conditioned Mixture Modeling.}
To explicitly model multimodal reasoning, we introduce a categorical strategy variable $c\in [M]$ representing up to $M$ distinct reasoning strategies. Rather than requiring $\mathbf z$ to represent all valid reasoning trajectories, $c$ partitions heterogeneous trajectories into distinct strategy modes, within which $\mathbf{z}$ captures trajectory-level variation. %Here, $M$ specifies the maximum number of reasoning strategies represented by the model. 
We sample a strategy once as $c\sim p_\theta(c\mid x)$ and keep it fixed throughout the latent trajectory, providing the conceptual hierarchy $(x,c)\rightarrow\mathbf z\rightarrow(\boldsymbol{\tau},y)$. Accordingly, we factorize the variational posterior to jointly infer the strategy and latent trajectory from the observed trace, \ie $q_\phi(c,\mathbf z\mid x,\boldsymbol{\tau}) = q_\phi(c\mid x,\boldsymbol{\tau})
\prod_{k=1}^{K}q_\phi(z_k\mid x,\boldsymbol{\tau}).$
The resulting ELBO of $\log p_\theta(y,\boldsymbol{\tau}\mid x)$ is given by modifying only the distribution alignment term in Eq.~\ref{eq:elbo} to incorporate the strategy $c$ as 
\begin{equation}\label{eq:objective}
\mathbb E_{q_\phi(\mathbf{z}\mid x,\boldsymbol{\tau})}[\log p_\theta(\boldsymbol{\tau}\mid\mathbf z)]
+
\mathbb E_{q_\phi(\mathbf{z}\mid x,\boldsymbol{\tau})}[\log p_\theta(y\mid\mathbf z,x)]
-
D_{\mathrm{KL}}\!\left(
q_\phi(c,\mathbf z\mid x,\boldsymbol{\tau})
\,\|\,p_\theta(c,\mathbf z\mid x)
\right).
\end{equation}
We use this ELBO as the training objective; details of its optimization is provided in Appendix~\ref{Appendix-subsec:Empirical Optimization}.

Following Proposition~\ref{pp:IB}, averaging the KL term in Eq.~\ref{eq:objective} over the data distribution $p(x,\boldsymbol{\tau})$ yields the aggregated alignment term $\mathbb E_x[D_{\mathrm{KL}}(\bar q_\phi(c,\mathbf z\mid x)\,\Vert{}\,p_\theta(c,\mathbf z\mid x))]$. 
Focusing on a given problem $x$ reveals why this parameterization naturally accommodates multimodal reasoning, as the point-wise KL divergence, $D_{\mathrm{KL}}(
\bar q_\phi(c,\mathbf z\mid x)
\,\|\,p_\theta(c,\mathbf z\mid x) )$, decomposes into
\begin{equation}
D_{\mathrm{KL}}\!\left(
\bar q_\phi(c\mid x)
\,\|\,p_\theta(c\mid x)
\right)
+
\mathbb E_{c\sim\bar q_\phi(c\mid x)}
\left[
D_{\mathrm{KL}}\!\left(
\bar q_\phi(\mathbf z\mid x,c)
\,\|\,p_\theta(\mathbf z\mid x,c)
\right)
\right],
\end{equation}
where $\bar q_\phi(\mathbf z\mid x,c)$ denotes the conditional distribution induced by the joint aggregated posterior. The strategy variable $c$ separates heterogeneous reasoning modes, so each conditional target $\bar q_\phi(\mathbf z\mid x,c)$ only needs to capture variation within a single strategy. When $c$ sufficiently separates the heterogeneous modes, each conditional target can be well approximated by a unimodal Gaussian, allowing the multimodal target distribution to be represented as a mixture of strategy-specific Gaussians.

% Using the same decomposition as in Proposition~\ref{pp:IB}, the KL term above decomposes, after averaging over $p(x,\boldsymbol{\tau})$, into a conditional mutual-information term and the distribution alignment objective $\mathbb E_x[D_{\mathrm{KL}}(\bar q_\phi(c,\mathbf z\mid x)\|p_\theta(c,\mathbf z\mid x))]$. Focusing on the latter, for each $x$ it further decomposes as
% \begin{align}
% &D_{\mathrm{KL}}\!\left(
% \bar q_\phi(c,\mathbf z\mid x)
% \,\|\,p_\theta(c,\mathbf z\mid x)
% \right)
% \nonumber\\
% &=
% D_{\mathrm{KL}}\!\left(
% \bar q_\phi(c\mid x)
% \,\|\,p_\theta(c\mid x)
% \right)
% +
% \mathbb E_{c\sim\bar q_\phi(c\mid x)}
% \left[
% D_{\mathrm{KL}}\!\left(
% \bar q_\phi(\mathbf z\mid x,c)
% \,\|\,p_\theta(\mathbf z\mid x,c)
% \right)
% \right]
% \end{align}
% where $\bar q_\phi(\mathbf z\mid x,c)$ denotes the conditional distribution induced by the joint aggregated posterior. By separating heterogeneous modes through $c$, each conditional target $\bar q_\phi(\mathbf z\mid x,c)$ only needs to capture within-strategy variation. When $c$ sufficiently separates heterogeneous reasoning modes, each conditional target can be well approximated by a single Gaussian.

\subsection{Implementing with Autoregressive Language Models}
\name builds on standard autoregressive language models and introduces lightweight modules for latent distribution prediction and latent injection, while leaving the backbone architecture unchanged. Implementation details are provided in Appendix~\ref{Appendix-sec:Implementation Details}.

\textbf{Shared Feature Transformation.}
At each latent step $k$, the language-model hidden representation $h_k$ is mapped into an intermediate representation $r_k = \operatorname{SwiGLU}(\operatorname{RMSNorm}(h_k))$~\citep{shazeer_SwiGLU_arXiv2020,zhang_RMSNorm_NeurIPS2019}. The posterior and prior share the backbone and nonlinear feature transformation up to $r_k$, while using separate linear heads to parameterize their respective distributions.

\textbf{Strategy-Conditioned Prior.} For each strategy $m\in[M]$, the prioir constructs a strategy-specific representation using Feature-wise Linear Modulation~(FiLM)~\citep{perez_FiLM_AAAI2018}, \ie $\tilde r_{m,k} = (1+\gamma_m)\odot r_k+\beta_m,$ where $(\gamma_m,\beta_m)$ are learnable strategy-specific parameters shared across all latent steps. At the first latent step, a linear layer followed by a softmax predicts the strategy distribution $p_\theta(c\mid x)$. Two linear heads then map each $\tilde r_{m,k}$ to the mean and log-variance of a $d$-dimensional Gaussian, $(\mu_{m,k}^p,\log((\sigma_{m,k}^p)^2))$, yielding
\begin{equation}
p_\theta(z_k\mid x,z_{<k},c=m)
=
\mathcal N\!\left(
\mu_{m,k}^p,
(\sigma_{m,k}^p)^2
\right),
\end{equation}
where $d$ can be chosen smaller than the language-model hidden dimension to encourage a compact latent reasoning space. After sampling $z_k$, a linear projection followed by LayerNorm maps it back to the language-model hidden dimension. The resulting vector is then injected as the continuous input embedding for the subsequent latent step.

\textbf{Parallel Target Latent Construction.}
To construct the target latent trajectory distribution in parallel, we append $K$ copies of a Latent-Reasoning special token, \texttt{[LR]}, to $(x,\boldsymbol{\tau})$. A single forward pass \textit{simultaneously} produces the hidden representation $h_k$ at the $k$-th \texttt{[LR]} position, which is transformed into $r_k$ using the shared feature transformation and then mapped by two linear heads to parameterize $q_\phi(z_k\mid x,\boldsymbol{\tau})=\mathcal N(\mu_k^q,(\sigma_k^q)^2)$.

\section{Related Works}\label{sec:rel}
Having discussed autoregressive language models, including conventional supervised fine-tuning and latent reasoning formulations, in Section~\ref{sec:desiderata}, we review here several additional lines of work that inform our approach.

\textbf{Joint-Embedding Predictive Architecture.} JEPA learns latent representations by predicting target embeddings rather than reconstructing every input detail, encouraging high-level semantic information~\citep{baevski_Data2vec_ICML2022,assran_JEPA_CVPR2023}. This predictive latent space has subsequently been exploited for planning and reasoning~\citep{maes_leworldmodel_ICML2025,liu_jepareasoner_arXiv2025}. However, these approaches operate on deterministic latent embeddings rather than explicitly modeling distributions over alternative trajectories. 

In particular, JEPA-Reasoner~\citep{liu_jepareasoner_arXiv2025} first learns a deterministic latent reasoning trajectory and then freezes the Reasoner while separately training a Talker to reconstruct the corresponding reasoning and answer sequence. Consequently, inference requires both autoregressive latent rollout and a subsequent verbalization stage, rather than predicting the answer directly from the latent trajectory. In contrast, \name learns a probabilistic latent trajectory distribution in an end-to-end manner and predicts the final answer directly from the latent trajectory without requiring CoT generation.

\textbf{Alternative Generative Paradigms for Latent Reasoning.}
A separate line of work redesigns the latent reasoning process itself: LaDiR~\citep{kang_LaDir_ICLR2026} performs reasoning through latent diffusion, while MarCos~\citep{liu_Marcos_arXiv2025} models reasoning as a hidden Markov chain of continuous thoughts. These methods first impose a latent space through auxiliary objectives on verbalized reasoning, such as VAE reconstruction with a Gaussian prior or sparsity regularization, and subsequently learn the corresponding latent reasoning dynamics. As a result, what information the latent representations retain is largely shaped by these predefined objectives rather than being jointly optimized for answer prediction and predictability from the problem. In contrast, \name jointly learns the target distributions and the problem-conditioned prior, allowing the latent reasoning distribution itself to adapt throughout training.

\section{Experiments}
\begin{table}[!t]
\centering

% =========================================================
% GPT-2 Results
% =========================================================
\begin{minipage}{\textwidth}
\centering
\scriptsize
\setlength{\tabcolsep}{1.8pt}
\renewcommand{\arraystretch}{1.08}

\captionof{table}{
Accuracy (\%) and token usage of GPT-2 across
in-distribution (ID) and out-of-distribution (OOD) tasks.
}
\label{tab:gpt2_results}

\begin{tabular}{@{}cl*{3}{cc}ccc|c@{}}
\toprule
\multirow{3}{*}{\textbf{Type}}
& \multirow{3}{*}{\textbf{Method}}
& \multicolumn{6}{c}{\textbf{ID Tasks}}
& \multicolumn{4}{c}{\textbf{OOD Tasks}} \\
\cmidrule(lr){3-8}
\cmidrule(lr){9-12}
&
& \multicolumn{2}{c}{\textbf{GSM8K-NL}}
& \multicolumn{2}{c}{\textbf{GSM8K}}
& \multicolumn{2}{c}{\textbf{CSQA}}
& \textbf{SVAMP}
& \textbf{GSM-Hard}
& \textbf{Multi-Arith}
& \multirow{2}{*}{
  \shortstack{\textbf{Avg.}\\\textbf{\# Tokens}\\\textbf{(OOD)}}
} \\
\cmidrule(lr){3-4}
\cmidrule(lr){5-6}
\cmidrule(lr){7-8}
\cmidrule(lr){9-9}
\cmidrule(lr){10-10}
\cmidrule(lr){11-11}
&
& \textbf{Acc.} & \textbf{\# Tokens}
& \textbf{Acc.} & \textbf{\# Tokens}
& \textbf{Acc.} & \textbf{\# Tokens}
& \textbf{Acc.}
& \textbf{Acc.}
& \textbf{Acc.}
& \\
\midrule
Verbalized
& CoT-SFT
& 30.93 & 67.00
& 39.88 & 31.50
& 34.81 & 67.60
& 34.20
& 6.70
& 81.80
& 42.60 \\
\midrule
\multirow{3}{*}{Deterministic}
& COCONUT
& 21.90 & 40.80
& 29.64 & 26.70
& 27.40 & 57.90
& 24.10
& 4.40
& 30.83
& 28.60 \\
& CODI
& 35.03 & 11.30
& \textbf{41.17} & 11.20
& 38.98 & 11.00
& 34.50
& 8.20
& 83.00
& 11.60 \\
& SIM-CoT
& 17.20 & 11.20
& 38.97 & 11.30
& 33.58 & 11.00
& 22.70
& 4.02
& 26.50
& 11.70 \\
\midrule
\multirow{3}{*}{Stochastic}
& ReGuLaR
& 19.32 & 6.30
& 17.89 & 6.40
& 20.10 & 12.40
& 22.38
& 3.79
& 31.87
& 5.70 \\
& CoLaR
& 20.32 & 33.20
& 36.39 & 17.30
& 21.62 & 16.50
& 25.70
& 4.09
& 54.17
& 24.60 \\
&
\cellcolor{blue!8}\textbf{PLRM (Ours)}
& \cellcolor{blue!8}\textbf{36.30}
& \cellcolor{blue!8}11.28
& \cellcolor{blue!8}38.40
& \cellcolor{blue!8}11.25
& \cellcolor{blue!8}\textbf{43.40}
& \cellcolor{blue!8}11.00
& \cellcolor{blue!8}\textbf{34.60}
& \cellcolor{blue!8}\textbf{8.26}
& \cellcolor{blue!8}\textbf{88.83}
& \cellcolor{blue!8}11.63 \\
\bottomrule
\end{tabular}
\end{minipage}

\vspace{1.0em}

% =========================================================
% Llama-1B Results
% =========================================================
\begin{minipage}{\textwidth}
\centering
\scriptsize
\setlength{\tabcolsep}{1.8pt}
\renewcommand{\arraystretch}{1.08}

\captionof{table}{
Accuracy (\%) and token usage of Llama-3.2-1B-Instruct across
in-distribution (ID) and out-of-distribution (OOD) tasks.
}
\label{tab:llama1b_results}

\begin{tabular}{@{}cl*{3}{cc}ccc|c@{}}
\toprule
\multirow{3}{*}{\textbf{Type}}
& \multirow{3}{*}{\textbf{Method}}
& \multicolumn{6}{c}{\textbf{ID Tasks}}
& \multicolumn{4}{c}{\textbf{OOD Tasks}} \\
\cmidrule(lr){3-8}
\cmidrule(lr){9-12}
&
& \multicolumn{2}{c}{\textbf{GSM8K-NL}}
& \multicolumn{2}{c}{\textbf{GSM8K}}
& \multicolumn{2}{c}{\textbf{CSQA}}
& \textbf{SVAMP}
& \textbf{GSM-Hard}
& \textbf{Multi-Arith}
& \multirow{2}{*}{
  \shortstack{\textbf{Avg.}\\\textbf{\# Tokens}\\\textbf{(OOD)}}
} \\
\cmidrule(lr){3-4}
\cmidrule(lr){5-6}
\cmidrule(lr){7-8}
\cmidrule(lr){9-9}
\cmidrule(lr){10-10}
\cmidrule(lr){11-11}
&
& \textbf{Acc.} & \textbf{\# Tokens}
& \textbf{Acc.} & \textbf{\# Tokens}
& \textbf{Acc.} & \textbf{\# Tokens}
& \textbf{Acc.}
& \textbf{Acc.}
& \textbf{Acc.}
& \\
\midrule
Verbalized
& CoT-SFT
& \textbf{53.22} & 74.05
& \textbf{58.91} & 29.67
& 63.39 & 75.00
& \textbf{58.60}
& 9.33
& 94.17
& 46.98 \\
\midrule
\multirow{3}{*}{Deterministic}
& COCONUT
& 33.13 & 45.50
& 48.93 & 27.58
& 70.84 & 10.35
& 50.60
& 7.96
& 83.83
& 33.41 \\
& CODI
& 42.76 & 13.21
& 54.44 & 13.19
& 67.32 & 12.00
& 56.00
& 9.40
& \textbf{94.83}
& 13.49 \\
& SIM-CoT
& 31.69 & 13.23
& 55.19 & 13.19
& 70.68 & 12.00
& 45.50
& 7.20
& 65.00
& 13.48 \\
\midrule
\multirow{3}{*}{Stochastic}
& ReGuLaR
& 18.60 & 7.46
& 35.38 & 7.67
& 59.60 & 8.14
& 40.76
& 3.91
& 35.43
& 6.95 \\
& CoLaR
& 29.34 & 37.07
& 44.35 & 17.17
& 42.51 & 23.63
& 39.40
& 6.75
& 75.33
& 28.20 \\
&
\cellcolor{blue!8}\textbf{PLRM (Ours)}
& \cellcolor{blue!8}42.08
& \cellcolor{blue!8}13.29
& \cellcolor{blue!8}50.00
& \cellcolor{blue!8}13.23
& \cellcolor{blue!8}\textbf{74.28}
& \cellcolor{blue!8}12.00
& \cellcolor{blue!8}49.40
& \cellcolor{blue!8}\textbf{10.01}
& \cellcolor{blue!8}\textbf{94.83} 
& \cellcolor{blue!8}13.62 \\
\bottomrule
\end{tabular}
\end{minipage}
\end{table}

\label{sec:experiments}
We conduct experiments on mathematical and commonsense reasoning tasks using GPT-2~\citep{radford_gpt2_openai2019} and Llama-3.2-1B-Instruct~\citep{grattafiori_llama1b_arxiv2024}, evaluating PLRM in both in-domain and out-of-domain settings. We further consider a multi-trajectory reasoning setting in which each problem admits multiple valid reasoning trajectories with corresponding answers.

\subsection{Experimental Setup}
\textbf{Datasets.} Following previous work~\citep{shen_Codi_EMNLP2025}, we train PLRM on three in-domain (ID) benchmarks: GSM8K-Aug~\citep{deng_GSM8K_arxiv2023}, its natural-language variant GSM8K-Aug-NL, and CommonsenseQA~\citep{talmor_commonsenseQA_NAACL2019}. For models trained on GSM8K-Aug-NL, we evaluate out-of-domain (OOD) generalization on three mathematical reasoning benchmarks: SVAMP~\citep{patel_SVAMP_NAACL2021}, GSM-Hard~\citep{gao_GsmHard_ICML2023}, and MultiArith~\citep{roy_MultiArith_EMNLP2015}. 

\noindent
\textbf{HiddenBranch Dataset.} To evaluate whether PLRM can capture heterogeneous valid reasoning trajectories, we construct HiddenBranch from the open-world abductive reasoning setting of ProofWriter~\citep{tafjord_proofwriter_ACL2021}. HiddenBranch asks the model to find a hypothesis that makes a target conclusion provable from the given facts and rules. For an input $x$, there may exist multiple valid reasoning trajectories with corresponding answers, $\{(\boldsymbol{\tau}_i, y_i)\}_{i=1}^{N_x}$, where each $\boldsymbol{\tau}_i$ includes a different valid hypothesis and the reasoning process induced by that hypothesis. Thus, the same input can admit multiple heterogeneous reasoning trajectories and answers.

\noindent
\textbf{Baselines.} We use verbalized CoT-SFT as a standard language-space reasoning baseline and compare PLRM with recent latent reasoning methods, which we group into deterministic and stochastic approaches. The deterministic baselines include COCONUT~\citep{hao_coconut_COLM2025}, CODI~\citep{shen_Codi_EMNLP2025}, and SIM-CoT~\citep{wei_SIMCoT_ICLR2026}, while the stochastic baselines include ReGuLaR~\citep{wang_regular_arXiv2026} and CoLaR~\citep{tan_CoLaR_NeurIPS2026}.

Further details on the datasets, the construction of \emph{HiddenBranch}, implementation details, and evaluation protocols are provided in Appendix~\ref{Appendix-sec:Experiment Details}.

\vspace{-1mm}

\subsection{Main Results}
\begin{wrapfigure}{r}{0.47\textwidth}
    \vspace{-1.0em}
    \centering
    \setlength{\intextsep}{6pt}
    \setlength{\columnsep}{10pt}
    
    \includegraphics[width=\linewidth]{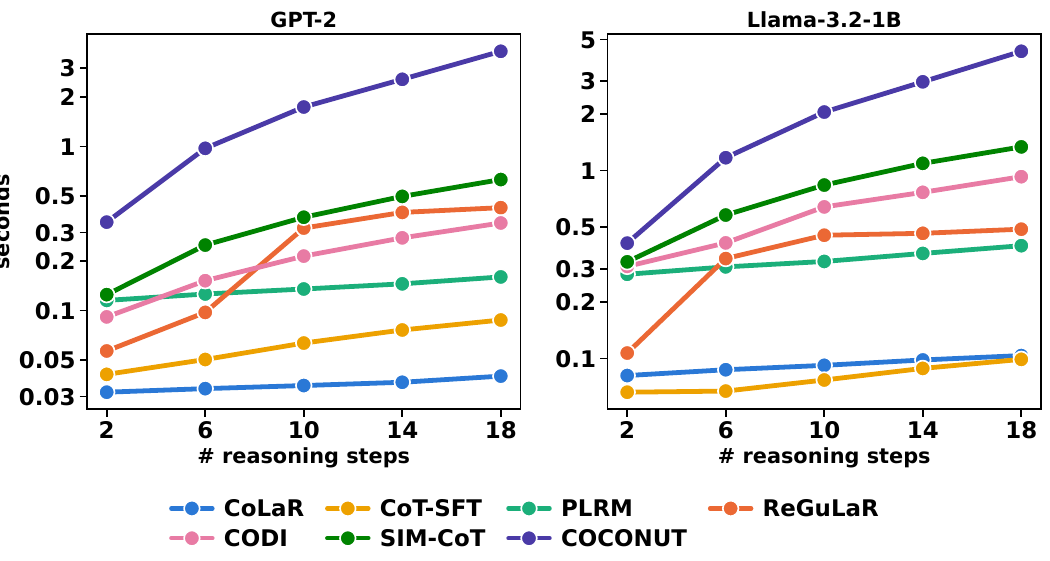}
    
    \caption{
        Average training time per iteration as the number of reasoning steps increases. For CoT-SFT, each step corresponds to four tokens, reflecting its approximately $4\times$ longer reasoning sequences.
    }
    \vspace{-1.0em}
    \label{fig:avg_step_sec}
\end{wrapfigure}

\textbf{Equation-Dominant vs Verbalized Reasoning.}
We distinguish two single-trajectory reasoning regimes according to how reasoning is expressed in the supervision. GSM8K-Aug primarily represents intermediate reasoning through mathematical expressions, such that most tokens directly encode the underlying computation. In contrast, the verbalized reasoning traces used for GSM8K-Aug-NL and CommonsenseQA contain richer natural-language explanations, where many tokens serve verbalization in addition to core reasoning.

This distinction is particularly important for latent reasoning. When the supervision is dominated by mathematical expressions, transferring the reasoning process into latent space mainly requires preserving the same computational content. In contrast, for verbalized reasoning traces, the model must compress a much longer sequence of $T$ tokens into only $K$ latent steps $(K \ll T)$, requiring it to retain core reasoning information while abstracting away trace-specific variation.
As shown in Tables~\ref{tab:gpt2_results} and~\ref{tab:llama1b_results}, \name remains competitive with deterministic latent reasoning methods on GSM8K-Aug, while showing particularly strong performance on GSM8K-Aug-NL and CommonsenseQA. Being trained on GSM8K-Aug-NL, \name also generalizes strongly to OOD mathematical reasoning tasks, achieving the best or tied-best accuracy in five of six model--task settings. This strong OOD generalization suggests that \name captures transferable reasoning rather than overfitting to specific patterns of verbalized CoT~\cite{shen_Codi_EMNLP2025}.

\textbf{Training Efficiency.}
Figure~\ref{fig:avg_step_sec} compares how the training cost scales with the number of reasoning steps. Methods trained with teacher forcing, including CoT-SFT, CoLaR, and \name, can process intermediate reasoning steps in parallel, resulting in only modest increases in training time as the number of latent steps grows. 
In contrast, all deterministic latent reasoning baselines (COCONUT, CODI, and SIM-CoT), together with the stochastic baseline (ReGuLaR), require sequential latent rollout during training, causing training cost to grow linearly with the number of latent steps. CoLaR is even faster because its target latent representations are constructed directly from summed token embeddings without an additional forward pass, but this efficiency comes at the cost of relying on heuristic targets and substantially lower empirical performance.

\begin{wraptable}{r}{0.36\textwidth}
\vspace{-0.8em}
\centering
\scriptsize
\setlength{\tabcolsep}{6pt}
\renewcommand{\arraystretch}{1.05}

\caption{Ablation study of PLRM.}
\label{tab:plrm_ablation}

\begin{tabular}{l|c}
    \toprule
    \textbf{Methods} & \textbf{Acc. (\%)} \\
    \midrule

    Fixed Posterior
    & 18.35\\

    \quad + Answer Loss
    & 19.49\\
\midrule
    CoLaR
    & 21.62\\

    \midrule

    \cellcolor{blue!8}\textbf{PLRM (Ours)}
    & \cellcolor{blue!8}\textbf{43.40} \\

    \bottomrule
\end{tabular}

\vspace{-0.8em}
\end{wraptable}

\textbf{Importance of Joint Posterior--Prior Training.}
We further examine whether jointly learning the posterior and prior is necessary.
Following a two-stage VAE setup~\cite{kang_LaDir_ICLR2026}, we first train the posterior on CommonsenseQA using GPT-2 backbone to reconstruct the verbalized reasoning trace while fixing the prior to a standard normal Gaussian.
We then extract latent trajectories for the entire training set, freeze the posterior, and train the problem-conditioned prior to predict these fixed targets.
We additionally consider a variant that includes answer prediction when learning the posterior, encouraging the latent representations to retain answer-relevant information. As shown in Table~\ref{tab:plrm_ablation}, the two-stage variants achieve only 18.35--19.49\% accuracy, comparable to CoLaR (21.62\%), suggesting a common limitation of learning from fixed latent targets.
In contrast, \name achieves 43.40\%, highlighting the importance of jointly adapting the target latent distribution and its problem-conditioned predictor.

\subsection{Heterogeneous Reasoning Strategies}
We next examine whether \name can model heterogeneous reasoning strategies when the same input admits multiple valid reasoning trajectories, as discussed in Section~\ref{subsec:Multimodal}. We assess how well the learned distribution captures these alternative hypotheses using \textbf{Hit Rate} and \textbf{Avg. TV Score}. Hit Rate measures whether greedy decoding recovers any valid answer. Avg. TV Score measures the agreement between the distribution of answers obtained from diverse reasoning trajectories and the distribution over all valid answers, jointly reflecting invalid generations and incomplete coverage of valid hypotheses; details are provided in Appendix~\ref{app:HB_metrics}.

\begin{wrapfigure}{r}{0.52\textwidth}
    \centering
    \includegraphics[width=\linewidth]{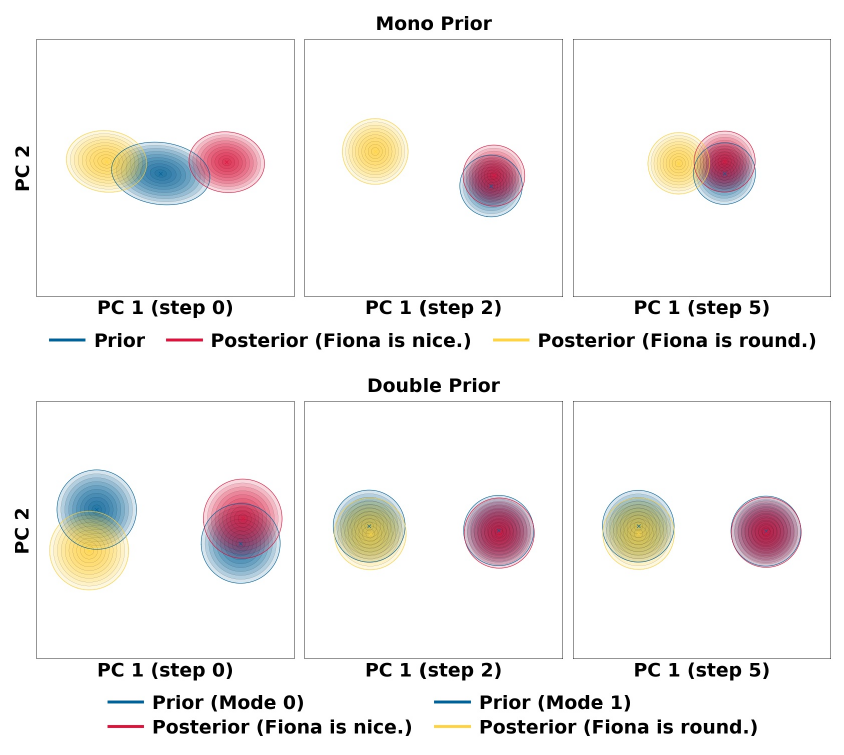}
    \caption{PCA visualization of prior and posterior latent distributions under mono~(M=1) and double-prior~(M=2) settings.}
    \label{fig:pca_2step}
\end{wrapfigure}

\textbf{Visualizing Multimodal Latent Distributions.}
We first compare the single-component prior in Eq.~\ref{eq:elbo} ($M=1$) with our mixture-based prior using two components ($M=2$). As shown in Figure~\ref{fig:pca_2step}, the Mono Prior initially expands its variance to cover distinct posterior modes. As reasoning proceeds, however, the posterior modes are progressively pulled together, collapsing the heterogeneous trajectories into a similar latent distribution. In contrast, the Double Prior assigns separate prior components to different posterior modes and maintains their separation across latent steps, preserving the multimodal structure without forcing heterogeneous reasoning trajectories to collapse. Additional visualizations for $M\geq3$ and qualitative examples reconstructing latent reasoning trajectories into verbalized reasoning traces are provided in Appendix~\ref{app:HB_analysis}.

\begin{wraptable}{r}{0.46\textwidth}
\vspace{-0.8em}
\centering
\begin{minipage}{\linewidth}
\centering
\scriptsize
\setlength{\tabcolsep}{4pt}
\renewcommand{\arraystretch}{1.1}

\caption{Hit rate and average TV Score across different methods.}
\label{tab:hit_tv_Score}

\begin{tabular}{lcccc}
    \toprule
    \textbf{Metric}
    & \textbf{CoT-SFT}& \textbf{ReGuLaR}
    & \textbf{CoLaR}
    & \cellcolor{blue!8}\textbf{PLRM} \\
    \midrule
    Hit Rate
    & 0.9116
    & 0.8665
    & 0.7397
    & \cellcolor{blue!8}\textbf{0.9205} \\
    Avg. TV Score& 0.6160
    & 0.4753
    & 0.3638
    & \cellcolor{blue!8}\textbf{0.7097} \\
    \bottomrule
\end{tabular}

\vspace{0.8em}

\caption{Ablation on the number of M.}
\label{tab:mode_ablation}

\begin{tabular}{lcccc}
    \toprule
    \textbf{Metric}
    & \textbf{M=1}& \cellcolor{blue!8}\textbf{M=2}& \textbf{M=3}& \textbf{M=4}\\
    \midrule
    Hit Rate
    & \textbf{0.9370}
    & \cellcolor{blue!8}0.9205
    & 0.9292
    & 0.9300 \\
    Avg. TV Score& 0.6252
    & \cellcolor{blue!8}\textbf{0.7097}
    & 0.6963
    & 0.6908 \\
    \bottomrule
\end{tabular}

\end{minipage}
\vspace{-0.8em}
\end{wraptable}

\textbf{Quantitative Evaluation.}
We compare \name with CoT-SFT and stochastic latent reasoning baselines, which can represent distributions over reasoning trajectories. As shown in Table~\ref{tab:hit_tv_Score}, it achieves the best performance on both Hit Rate and Avg. TV Score, improving the latter from 0.6160 to 0.7097 over CoT-SFT, a 15.2\% relative gain that indicates better coverage of heterogeneous valid hypotheses while maintaining high deterministic accuracy.
Table~\ref{tab:mode_ablation} examines the number of mixture components.
While a single mode achieves high Hit Rate, its substantially lower TV Score indicates limited coverage of alternative valid hypotheses. Using two modes yields the best TV Score, indicating that the heterogeneity in HiddenBranch can be sufficiently modeled with two strategy components.
\section{Conclusion}
\label{sec:conclusion}

In this work, we proposed \name, a variational framework for probabilistic latent reasoning with dense supervision and parallel teacher-forced training.
By jointly learning posterior target distributions and a problem-conditioned prior, \name captures core reasoning information while abstracting away trace-specific verbalization, and models heterogeneous strategies through a mixture-based prior.
Across mathematical and commonsense reasoning tasks, \name outperforms existing stochastic latent reasoning methods while remaining competitive with deterministic approaches.
We envision \name as a step toward scalable probabilistic reasoning in continuous latent space beyond verbalized chain-of-thought.

\clearpage
% \subsection*{AI use statement}
% In this work, we used generative Al tools to aid and polish writing. We did not use generative AI tools for any other tasks requiring disclosure under the ICLR policy. We reviewed all AI-assisted content and take full responsibility for the final content of this work.

% \subsection*{Ethics Statement}
% This work uses publicly available datasets and involves no direct interaction with human subjects. We additionally construct HiddenBranch by transforming examples from the existing ProofWriter dataset to create multiple valid reasoning trajectories for the same input. We plan to publicly release the resulting dataset and its construction procedure to support reproducibility and future research.

% \subsection*{Reproducibility Statement}
% We will publicly release all code and datasets used in this work to facilitate reproducibility.


\bibliographystyle{iclr2027_conference}
\bibliography{iclr2027_conference}

\appendix
\newpage

\section{Implementation Details}
\label{Appendix-sec:Implementation Details}

\begin{figure}[h]
    \centering
    \includegraphics[width=0.7\linewidth]{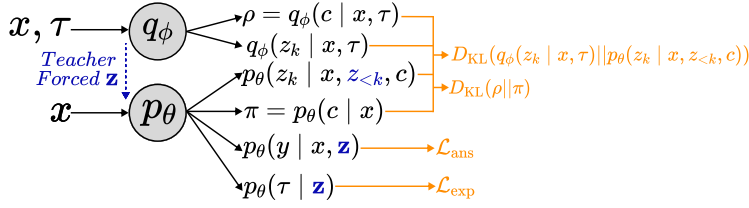}
    \caption{Overall Training Computation}
    \label{fig:overall_training}
\end{figure}

\subsection{Empirical Optimization}
\label{Appendix-subsec:Empirical Optimization}

The objective in Eq.~\eqref{eq:objective} contains the following trajectory-level KL divergence:
\begin{equation}
D_{\mathrm{KL}}\!\left(
q_\phi(c,\mathbf{z}\mid x,\boldsymbol{\tau})
\,\|\, 
p_\theta(c,\mathbf{z}\mid x)
\right).
\end{equation}
Using the factorization of the posterior and prior, this term can be
decomposed as
\begin{align}
&D_{\mathrm{KL}}\!\left(
q_\phi(c,\mathbf{z}\mid x,\boldsymbol{\tau})
\,\|\,
p_\theta(c,\mathbf{z}\mid x)
\right)
\nonumber\\
&=
D_{\mathrm{KL}}\!\left(
q_\phi(c\mid x,\boldsymbol{\tau})
\,\|\,
p_\theta(c\mid x)
\right)
\nonumber\\
&\quad+
\mathbb{E}_{q_\phi(c\mid x,\boldsymbol{\tau})}
\sum_{k=1}^{K}
\mathbb{E}_{q_\phi(z_{<k}\mid x,\boldsymbol{\tau})}
\left[
D_{\mathrm{KL}}\!\left(
q_\phi(z_k\mid x,\boldsymbol{\tau})
\,\|\,
p_\theta(z_k\mid x,z_{<k},c)
\right)
\right].
\label{eq:strategy_kl}
\end{align}

The expectation over $z_{<k}$ is estimated by Monte Carlo samples from the
posterior. We therefore focus on the inner KL term for a fixed strategy
$c=m$. Both the posterior and prior are parameterized as diagonal Gaussian
distributions:
\begin{align}
q_\phi(z_k\mid x,\boldsymbol{\tau})
&=
\mathcal{N}\!\left(
\mu_q^k,
\operatorname{diag}\!\left((\sigma_q^k)^2\right)
\right),\\
p_\theta(z_k\mid x,z_{<k},c=m)
&=
\mathcal{N}\!\left(
\mu_{p,m}^k,
\operatorname{diag}\!\left((\sigma_{p,m}^k)^2\right)
\right).
\end{align}

For two $d$-dimensional Gaussian distributions,
$q=\mathcal{N}(\mu_q,\Sigma_q)$ and
$p=\mathcal{N}(\mu_p,\Sigma_p)$, the KL divergence is
\begin{equation}
D_{\mathrm{KL}}(q\|p)
=
\frac{1}{2}
\left[
\log\frac{|\Sigma_p|}{|\Sigma_q|}
-d
+
\operatorname{tr}\!\left(\Sigma_p^{-1}\Sigma_q\right)
+
(\mu_p-\mu_q)^\top
\Sigma_p^{-1}
(\mu_p-\mu_q)
\right].
\end{equation}
Since both covariance matrices are diagonal,
\begin{align}
\log\frac{|\Sigma_p|}{|\Sigma_q|}
&=
\sum_{j=1}^{d}
\log
\frac{(\sigma_{p,m,j}^k)^2}
     {(\sigma_{q,j}^k)^2},\\
\operatorname{tr}\!\left(\Sigma_p^{-1}\Sigma_q\right)
&=
\sum_{j=1}^{d}
\frac{(\sigma_{q,j}^k)^2}
     {(\sigma_{p,m,j}^k)^2},\\
(\mu_p-\mu_q)^\top
\Sigma_p^{-1}
(\mu_p-\mu_q)
&=
\sum_{j=1}^{d}
\frac{
(\mu_{q,j}^k-\mu_{p,m,j}^k)^2
}{
(\sigma_{p,m,j}^k)^2
}.
\end{align}
Substituting these terms gives the closed-form Gaussian KL at latent step
$k$ for strategy $m$:
\begin{equation}
D_{\mathrm{KL}}\!\left(
q_\phi
\,\|\,
p_\theta
\right)
=
\frac{1}{2}
\sum_{j=1}^{d}
\left[
\log
\frac{(\sigma_{p,m,j}^k)^2}
     {(\sigma_{q,j}^k)^2}
+
\frac{
(\sigma_{q,j}^k)^2
+
(\mu_{q,j}^k-\mu_{p,m,j}^k)^2
}{
(\sigma_{p,m,j}^k)^2
}
-1
\right].
\label{eq:closed_form_gaussian_kl}
\end{equation}

Using this closed form, the expectation over the discrete strategy variable
in Eq.~\eqref{eq:strategy_kl} can be written as a
responsibility-weighted sum:
\begin{align}
&\mathbb{E}_{q_\phi(c\mid x,\boldsymbol{\tau})}
\sum_{k=1}^{K}
\mathbb{E}_{q_\phi(z_{<k}\mid x,\boldsymbol{\tau})}
\left[
D_{\mathrm{KL}}\!\left(
q_\phi(z_k\mid x,\boldsymbol{\tau})
\,\|\,
p_\theta(z_k\mid x,z_{<k},c)
\right)
\right]
\nonumber\\
&=
\sum_{k=1}^{K}
\mathbb{E}_{q_\phi(z_{<k}\mid x,\boldsymbol{\tau})}
\left[
\sum_{m=1}^{M}
\rho_m
D_{\mathrm{KL}}\!\left(
q_\phi(z_k\mid x,\boldsymbol{\tau})
\,\|\,
p_\theta(z_k\mid x,z_{<k},c=m)
\right)
\right],
\label{eq:responsibility_weighted_kl}
\end{align}
where
$\rho_m=q_\phi(c=m\mid x,\boldsymbol{\tau})$.
The remaining expectation over $z_{<k}$ is approximated using posterior
Monte Carlo samples.

\subsection{Practical Details of PLRM}
\label{Appendix-subsec:Practical Details of PLRM}
Although the formulation in Section~\ref{sec:problem} specifies the main training objectives of PLRM, we employ several additional regularization and optimization techniques to stabilize the joint training of the posterior and mixture prior. For brevity, throughout this subsection we denote
\[
q_k \equiv q_\phi(z_k \mid x,\boldsymbol{\tau}), \qquad
p_{k,m} \equiv p_\theta(z_k \mid x,z_{<k},c=m),
\]
and let $\rho_m \equiv q_\phi(c=m\mid x,\boldsymbol{\tau})$ denote the EM-style responsibility defined from the first latent step.

\textbf{EM-style Strategy Learning.} Inspired by divergence-based soft clustering~\citep{banerjee_CBD_JMLR2005}, we use an EM-style soft assignment to stabilize strategy learning. Rather than introducing a separate classifier for $c$, we perform an E-step:
\begin{equation}
\rho_m
\propto
\exp\!\left(
-\frac{1}{T}
D_{\mathrm{KL}}\!\left(
q_\phi(z_1\mid x,\boldsymbol{\tau})
\,\|\,p_\theta(z_1\mid x,c=m)
\right)
\right),
\end{equation}
where $T$ controls the assignment sharpness~\citep{ueda_DAEM_NN1998}. The resulting responsibilities are fixed across the latent trajectory and used to update the corresponding strategy-specific prior components and $p_\theta(c\mid x)$, analogous to a gradient-based M-step. We infer the strategy from the first latent distribution because it is the only prior prediction made before any teacher-forced posterior latent is available. This restriction does not limit the intended multimodal modeling, since distinct reasoning strategies can already induce different modes at $z_1$, as shown above.

\paragraph{Mixture-Preserving KL Regularization.}
A direct responsibility-weighted Gaussian alignment, $\sum_m \rho_m D_{\mathrm{KL}}(q_k \| p_{k,m})$,
encourages multiple prior components assigned to the same posterior $q_k$ to move toward that common posterior target, thereby reducing the separation between mixture components. To isolate this contraction effect, we adopt the exponential-family decomposition derived in DEM-VAE~\citep{shi_DEMVAE_ICLR2020}:
\begin{equation}
    \sum_m \rho_m D_{\mathrm{KL}}(q_k \| p_{k,m})
    =
        D_{\mathrm{KL}}(q_k \| p_{\bar{\eta}_k})
    +
    L_{d,k},
\end{equation}
where $\bar{\eta}_k=\sum_m \rho_m \eta_{m,k}$ is the responsibility-weighted average of the natural parameters of $p_{k,m}$, and $p_{\bar{\eta}_k}$ denotes the corresponding Gaussian distribution. The dispersion term $L_{d,k}$ is given by
\begin{equation}
    L_{d,k}
    =
    \sum_m \rho_m A(\eta_{m,k})
    -
    A(\bar{\eta}_k)
    \geq 0,
\end{equation}
where $A(\cdot)$ denotes the Gaussian log-partition function. The non-negative term $L_{d,k}$ captures the additional cost induced by dispersion among the responsibility-weighted prior components and vanishes when these components coincide. Consequently, minimizing the original responsibility-weighted KL also penalizes separation among the prior components, introducing pressure toward component collapse. To attenuate this effect, we optimize the corrected Gaussian alignment
\begin{equation}
    \mathcal{L}_{\mathrm{GKL},k}
    =
    \sum_m \rho_m D_{\mathrm{KL}}(q_k \| p_{k,m})
    -
    \gamma L_{d,k},
\end{equation}
which reduces the component-collapse pressure induced by responsibility-weighted KL alignment while retaining a non-negative distribution-matching objective for $\gamma \in [0,1]$.

Following the KL-balancing strategy of DreamerV2~\citep{hafner_dreamerv2_iclr2021}, we further stabilize training by separately controlling the gradients flowing to the predictive prior and target posterior. For each component-wise Gaussian KL term, we use
\begin{equation}
    0.8\,
    D_{\mathrm{KL}}\!\left(
        \operatorname{sg}(q_k)\,\|\,p_{k,m}
    \right)
    +
    0.2\,
    D_{\mathrm{KL}}\!\left(
        q_k\,\|\,\operatorname{sg}(p_{k,m})
    \right),
\end{equation}
where $\operatorname{sg}(\cdot)$ denotes the stop-gradient operator. This places greater optimization pressure on the predictive prior to follow the posterior target.

\paragraph{Posterior Representation Regularization.}
As described in the main text, we apply VISReg to the posterior means rather than sampled latent variables. Concretely, VISReg consists of three complementary terms. The \emph{center} term keeps the batch-aggregated
posterior means close to zero, the \emph{scale} term encourages each latent dimension to maintain unit batch-wise standard deviation, and the \emph{shape} term encourages the standardized posterior means to follow a
standard Gaussian under random one-dimensional projections. The resulting regularizer is
\begin{equation}
    \mathcal{L}_{\mathrm{VISReg}}
    =
    \mathcal{L}_{\mathrm{scale}}
    +
    \mathcal{L}_{\mathrm{shape}}
    +
    \mathcal{L}_{\mathrm{center}}.
\end{equation}
For the scale term, we use a squared deviation from unit standard deviation, $(1-\sigma)^2$, rather than a hinge penalty, so that the regularizer continues to provide a gradient when the batch standard deviation exceeds the target value. For the shape term, standardized posterior means are projected onto random unit directions, sorted across the batch, and matched to the corresponding quantiles of a standard Gaussian.

% \paragraph{Strategy-Usage Regularization.}
% The mixture prior can degenerate if most trajectories are assigned to a single strategy component. To discourage such global mode collapse, we regularize the marginal strategy usage. Let
% \begin{equation}
%     \bar{\rho}
%     =
%     \frac{1}{N}\sum_{n=1}^{N}\rho_n
% \end{equation}
% denote the batch-aggregated responsibility distribution. We minimize
% \begin{equation}
%     \mathcal{L}_{\mathrm{bal}}
%     =
%     D_{\mathrm{KL}}\!\left(
%         \bar{\rho}\,\|\,\mathrm{Uniform}(M)
%     \right),
% \end{equation}
% which encourages the strategy components to be utilized across the training distribution. We do not directly penalize the entropy of each individual responsibility vector; its conditional entropy is used only as a diagnostic for monitoring assignment sharpness.

\paragraph{Overall Training Objective.}
With the dispersion correction, the practical distribution-alignment term is
\begin{equation}
    \mathcal{L}_{\mathrm{align}}
    =
    D_{\mathrm{KL}}\!\left(
        \rho \,\|\, p_\theta(c\mid x)
    \right)
    +
    \frac{1}{K}
    \sum_{k=1}^{K}
    \left[
        \sum_m \rho_m
        D_{\mathrm{KL}}(q_k \| p_{k,m})
        -
        \gamma L_{d,k}
    \right],
\end{equation}
where the categorical KL is applied once per trajectory, whereas the Gaussian alignment is averaged over the $K$ latent steps. The resulting training objective is
\begin{equation}
    \mathcal{L}
    =
    \mathcal{L}_{\mathrm{ans}}
    +
    \mathcal{L}_{\mathrm{exp}}
    +
    \beta(1-\lambda_{\mathrm{reg}})
    \mathcal{L}_{\mathrm{align}}
    +
    \beta\lambda_{\mathrm{reg}}
    \mathcal{L}_{\mathrm{VISReg}}.
\end{equation}
We linearly warm up the shared coefficient $\beta$ during training. Detailed values of the loss coefficients and training hyperparameters are provided in the following subsection.

\subsection{Method Configuration}
\label{Appendix-subsec:Method Configuration}
Table~\ref{tab:plrm_configuration} summarizes the main optimization and method-specific hyperparameters used for PLRM. Unless otherwise specified, we use $K=6$ latent reasoning steps and optimize all models with AdamW using cosine learning-rate decay and BF16 mixed precision. For the main benchmarks, including GSM8K, GSM8K-NL, and CommonsenseQA, we use a single prior component ($M=1$), as each example provides a single reasoning trajectory. Dataset- and backbone-specific settings, including the learning rate, number of epochs, batch size, posterior context allocation, and objective coefficients, are reported in the table.

\begin{table*}[h]
\centering
\caption{
Main training and method-specific hyperparameters of PLRM.
PTL denotes the number of posterior-visible CoT tokens per latent step.
}
\label{tab:plrm_configuration}
\resizebox{\textwidth}{!}{
\begin{tabular}{lcccccc}
\toprule
& \multicolumn{3}{c}{\textbf{GPT-2}}
& \multicolumn{3}{c}{\textbf{Llama-3.2-1B}} \\
\cmidrule(lr){2-4}
\cmidrule(lr){5-7}
\textbf{Hyperparameter}
& \textbf{GSM8K}
& \textbf{GSM8K-NL}
& \textbf{CQA}
& \textbf{GSM8K}
& \textbf{GSM8K-NL}
& \textbf{CQA} \\
\midrule

\multicolumn{7}{l}{\textit{Optimization}} \\

Learning rate
& $1\times10^{-3}$
& $1\times10^{-3}$
& $2\times10^{-3}$
& $3\times10^{-4}$
& $3\times10^{-4}$
& $3\times10^{-4}$ \\

Epochs
& 40
& 40
& 40
& 40
& 40
& 10 \\

Effective batch size
& 128
& 128
& 128
& 256
& 256
& 256 \\

\midrule
\multicolumn{7}{l}{\textit{Latent Configuration}} \\

Posterior tokens per latent (PTL)
& 8
& 12
& 24
& 7
& 14
& 24 \\

Max.\ input tokens
& 200
& 256
& 1000
& 200
& 256
& 1000 \\

\midrule
\multicolumn{7}{l}{\textit{Objective and Regularization}} \\

KL coefficient $\beta_{\max}$
& 3.0
& 3.0
& 3.0
& 3.0
& 3.0
& 3.0\\

KL warmup ratio
& 0.8
& 0.8
& 0.8
& 0.8
& 0.8
& 0.8 \\

KL balance $\alpha$
& 0.8
& 0.8
& 0.8
& 0.8
& 0.8
& 0.8 \\

VISReg weight $\lambda_{\mathrm{reg}}$
& 0.9
& 0.9
& 0.9
& 0.9
& 0.9
& 0.9 \\

VISReg projections
& 512
& 512
& 512
& 512
& 512
& 512 \\

\bottomrule
\end{tabular}
}
\end{table*}
Notably, we use the same KL- and VISReg-related hyperparameters across
different datasets and backbone architectures without per-setting tuning.
Their consistent effectiveness across these settings suggests that PLRM is
not highly sensitive to these regularization choices. In contrast, the posterior-token allocation per latent step (PTL) is adjusted
by dataset to account for differences in reasoning-trace length, such that
later latent targets are conditioned on progressively larger portions of the
CoT trajectory. We fix the Gaussian latent dimension to $d=768$ across all datasets and backbone architectures.

%%%%%%%%%%%%%%%%%%%%%%%%%%%%%%%%%%%%%%%%%
\newpage
\section{EXPERIMENTAL DETAILS}
\label{Appendix-sec:Experiment Details}

\subsection{Benchmark Datasets}
\label{Appendix-subsec:Benchmark Datasets}

\paragraph{In-Distribution Datasets.}
Following prior latent reasoning work~\citep{shen_Codi_EMNLP2025}, we consider three in-distribution (ID) benchmarks spanning mathematical and commonsense reasoning.

\textbf{GSM8K-Aug.}
GSM8K-Aug~\citep{deng_GSM8K_arxiv2023} is an augmented version of the GSM8K mathematical reasoning benchmark. It consists of grade-school-level arithmetic word problems paired with intermediate reasoning traces and final answers. We use this dataset to evaluate latent reasoning in a standard mathematical reasoning setting.

\textbf{GSM8K-Aug-NL.}
GSM8K-Aug-NL is the natural-language variant of GSM8K-Aug introduced in the same experimental setting~\citep{deng_GSM8K_arxiv2023}. Compared with the standard GSM8K-Aug formulation, it provides reasoning trajectories in a more natural-language-oriented form. We use GSM8K-Aug-NL as an additional mathematical reasoning benchmark and use models trained on this dataset for our out-of-distribution evaluation.

\textbf{CommonsenseQA.}
CommonsenseQA (CSQA)~\citep{talmor_commonsenseQA_NAACL2019} is a multiple-choice question answering benchmark designed to evaluate commonsense reasoning. Unlike the GSM8K-based datasets, which primarily require arithmetic reasoning, CSQA requires reasoning over implicit commonsense knowledge. We include CSQA to evaluate whether PLRM is effective beyond mathematical reasoning.

\paragraph{Out-of-Distribution Datasets.}
To evaluate out-of-distribution (OOD) generalization, we evaluate models trained on GSM8K-Aug-NL on three mathematical reasoning benchmarks that are not used during training.

\textbf{SVAMP.}
SVAMP~\citep{patel_SVAMP_NAACL2021} is a collection of arithmetic word problems designed to test robustness to variations in problem structure and linguistic formulation. The benchmark provides a distribution shift from the GSM8K-based training data while retaining the requirement for multi-step arithmetic reasoning.

\textbf{GSM-Hard.}
GSM-Hard~\citep{gao_GsmHard_ICML2023} is a more challenging variant of GSM8K constructed by replacing numerical values with larger and less common numbers. This modification preserves the underlying reasoning structure while increasing the difficulty of numerical computation, allowing us to evaluate robustness under a controlled distribution shift.

\textbf{MultiArith.}
MultiArith~\citep{roy_MultiArith_EMNLP2015} consists of multi-step arithmetic word problems that require composing several arithmetic operations to obtain the final answer. We use MultiArith to assess whether reasoning learned from GSM8K-Aug-NL transfers to a distinct mathematical word-problem distribution.

\subsection{\emph{HiddenBranch} Dataset}
\label{Appendix-subsec:HiddenBranch Dataset}
\paragraph{HiddenBranch Construction.}
We construct \emph{HiddenBranch} from the open-world abductive reasoning setting of ProofWriter~\citep{tafjord_proofwriter_ACL2021}, using examples from depths 1, 2, 3, and 5. We retain source questions that admit between two and five valid hidden facts, where each hidden fact independently makes the query provable when added to the original facts and rules. For each hidden fact, we use the corresponding ProofWriter proof as the ground-truth reasoning trajectory after verifying that the hidden fact independently
entails the query.

A key property of \emph{HiddenBranch} is that a single unique input can be associated with multiple valid reasoning--answer pairs. The input, consisting of the task instruction, facts, rules, and question, remains identical across branches, while each branch provides a different ground-truth reasoning trajectory and corresponding answer. During training and validation, these branches are represented as separate supervised examples sharing the same input. At test time, each unique input is presented only once and evaluated against the complete set of valid hidden answers.

\paragraph{Example.}
The following example illustrates two branches associated with the same unique input.\\

\begin{quote}
\small
\textbf{Input}

\textbf{Task:}
Find one missing fact that, when assumed by itself in addition to the given Facts and Rules, makes the Question provable. Give one proof for that missing fact.

\textbf{Facts:}

\texttt{triple3. Harry is green.}

\texttt{triple2. Dave is rough.}

\texttt{triple1. Anne is cold.}

\textbf{Rules:}

\texttt{rule1. All cold, round people are green.}

\texttt{rule2. If Harry is round and Harry is not rough then Harry is big.}

\texttt{rule3. If Anne is young and Anne is rough then Anne is big.}

\texttt{rule4. All green people are young.}

\texttt{rule5. Young, big people are kind.}

\texttt{rule6. If someone is cold then they are round.}

\textbf{Question:}

\texttt{Anne is kind.}
\end{quote}

For this same input, \emph{HiddenBranch} contains multiple valid ground-truth reasoning--answer pairs:

\begin{quote}
\small
\textbf{GT Pair 1}

\textbf{GT-CoT:}

\texttt{B1: Assume Anne is big.}

\texttt{Anne is cold. --rule6--> Anne is round.;}

\texttt{Anne is cold. \& Anne is round. --rule1--> Anne is green.;}

\texttt{Anne is green. --rule4--> Anne is young.;}

\texttt{Anne is young. \& Anne is big. --rule5--> Anne is kind.}

\textbf{GT-Answer:}

\texttt{["Anne is big."]}

\vspace{0.8em}

\textbf{GT Pair 2}

\textbf{GT-CoT:}

\texttt{B1: Assume Anne is rough.}

\texttt{Anne is cold. --rule6--> Anne is round.;}

\texttt{Anne is cold. \& Anne is round. --rule1--> Anne is green.;}

\texttt{Anne is green. --rule4--> Anne is young.;}

\texttt{Anne is young. \& Anne is rough. --rule3--> Anne is big.;}

\texttt{Anne is young. \& Anne is big. --rule5--> Anne is kind.}

\textbf{GT-Answer:}

\texttt{["Anne is rough."]}
\end{quote}

Although the two examples share exactly the same input, they correspond to different valid hidden facts and therefore induce different reasoning trajectories and answers. This construction explicitly creates a one-to-many mapping from a unique input to multiple valid reasoning--answer pairs, which allows us to evaluate whether a stochastic latent reasoning model can represent and recover heterogeneous reasoning strategies for the same problem.

\begin{table}[h]
\centering
\small
\setlength{\tabcolsep}{6pt}
\caption{
Statistics of the \emph{HiddenBranch} dataset.
Each unique input corresponds to one source question, while valid branches
denote distinct reasoning--answer pairs associated with that input.
}
\label{tab:hiddenbranch_stats}
\begin{tabular}{lrr}
\toprule
\textbf{Split}
& \textbf{\# Unique Inputs}
& \textbf{\# Valid Branches} \\
\midrule
Train      & 21,466 & 52,734 \\
Validation & 3,179  & 7,712 \\
Test       & 6,128  & 15,101 \\
\midrule
Total      & 30,773 & 75,547 \\
\bottomrule
\end{tabular}
\end{table}

\paragraph{Evaluation Metrics.}~\label{app:HB_metrics}
For the multi-trajectory setting, we evaluate both deterministic answer correctness and the distribution of stochastic generations. Let $G_i$ denote the set of valid gold answers for input $i$.

\textbf{Hit Rate.}
We first perform greedy decoding and measure whether the deterministically generated answer belongs to the set of valid answers:
\begin{equation}
    \mathrm{HitRate}
    =
    \frac{1}{N}
    \sum_{i=1}^{N}
    \mathbf{1}
    \left[
        \hat{a}^{\mathrm{det}}_i \in G_i
    \right].
\end{equation}
Thus, Hit Rate measures whether the model can recover at least one valid reasoning branch under deterministic decoding.

\textbf{TV Score.}
To evaluate stochastic generations, we draw $R$ samples for each input and compare their empirical answer distribution with an ideal distribution that assigns equal probability to all valid gold answers and zero probability to invalid generations. For an input with $|G_i|=G$ gold answers, let $n_{i,g}$ be the number of samples matching gold answer $g$, and let $n_{i,\mathrm{inv}}=R-\sum_{g=1}^{G} n_{i,g}$ denote the number of invalid samples. We define
\begin{equation}
    P_i
    =
    \left[
        \frac{n_{i,1}}{R},\ldots,
        \frac{n_{i,G}}{R},
        \frac{n_{i,\mathrm{inv}}}{R}
    \right],
    \qquad
    Q_i
    =
    \left[
        \frac{1}{G},\ldots,
        \frac{1}{G},
        0
    \right].
\end{equation}
The per-example score is the complement of the total variation distance,
\begin{equation}
    S_i
    =
    1-\mathrm{TV}(P_i,Q_i)
    =
    1-\frac{1}{2}\sum_c |P_i(c)-Q_i(c)|,
\end{equation}
and we report
\begin{equation}
    \mathrm{TVScore}
    =
    \frac{1}{N}\sum_{i=1}^{N} S_i.
\end{equation}
The score ranges from $0$ to $1$, with higher values indicating better agreement with the ideal distribution. It jointly penalizes invalid generations and concentration on only a subset of valid branches: the score is maximized when all generations are valid and are distributed as uniformly as possible across the gold answers.

\newpage

\section{ADDITIONAL EXPERIMENTAL RESULTS}
Overview. In this section, we provide additional experimental results and analyses of PLRM. We first present a thorough comparison with previously reported results on in-distribution benchmarks, covering verbalized, deterministic latent, and stochastic latent reasoning methods across both GPT-2 and Llama-3.2-1B-Instruct backbones. We then focus on the multi-trajectory setting and analyze how the learned prior captures multiple trajectory-specific posterior modes as the number of valid reasoning branches increases. Finally, we provide qualitative examples showing that individual strategy modes induce distinct valid reasoning trajectories and corresponding answers.

\subsection{Comparison with Previously Reported Results}
 \begin{table}[h]
\centering
\scriptsize
\setlength{\tabcolsep}{3.8pt}
\renewcommand{\arraystretch}{1.08}
\caption{
Accuracy (\%) of GPT-2 and Llama-3.2-1B-Instruct on in-distribution tasks.
}
\label{tab:appendix_id_paper_performance}
\begin{tabular}{clccc|ccc}
\toprule
\multirow{2}{*}{\textbf{Type}}
& \multirow{2}{*}{\textbf{Method}}
& \multicolumn{3}{c|}{\textbf{GPT-2}}
& \multicolumn{3}{c}{\textbf{Llama-3.2-1B}} \\
\cmidrule(lr){3-5}
\cmidrule(lr){6-8}
&
& \textbf{GSM8K-NL}
& \textbf{GSM8K}
& \textbf{CSQA}
& \textbf{GSM8K-NL}
& \textbf{GSM8K}
& \textbf{CSQA} \\
\midrule

Verbalized
& Explicit CoT
& 34.8
& \textbf{44.1}
& 36.9
& \textbf{54.1}
& \textbf{61.6}
& 68.2 \\
\midrule

\multirow{3}{*}{Deterministic}
& COCONUT
& --
& 34.1
& --
& --
& --
& -- \\
& CODI
& 35.3
& 43.7
& \textbf{44.0}
& 49.7
& 55.6
& 74.0 \\
& SIM-CoT
& --
& 42.6
& --
& --
& 56.1
& -- \\
\midrule

\multirow{2}{*}{Stochastic}
& ReGuLaR
& --
& --
& --
& 20.2
& 34.9
& -- \\
& CoLaR
& --
& --
& --
& --
& 40.1
& -- \\
&
\cellcolor{blue!8}\textbf{PLRM (Ours)}
& \cellcolor{blue!8}\textbf{36.30}
& \cellcolor{blue!8}38.40
& \cellcolor{blue!8}43.40
& \cellcolor{blue!8}42.08
& \cellcolor{blue!8}50.00
& \cellcolor{blue!8}\textbf{74.28} \\
\bottomrule
\end{tabular}
\end{table}
Table~\ref{tab:appendix_id_paper_performance} complements our main in-distribution evaluation with performance reported in prior work. Overall, the comparison is consistent with the distinction between equation-dominant and verbalized reasoning discussed in Section~5.2. On GSM8K, where intermediate reasoning is largely represented through mathematical operations, PLRM remains competitive with deterministic latent reasoning methods. In contrast, PLRM shows stronger performance on tasks with more heavily verbalized reasoning traces, where the model must compress lengthy natural-language reasoning into a small number of latent steps while preserving the information necessary for answer prediction. Importantly, across all settings where previously reported stochastic baselines are available, PLRM consistently achieves higher accuracy than existing stochastic latent reasoning methods. This suggests that PLRM improves the effectiveness of stochastic latent reasoning without sacrificing the competitive performance observed in deterministic latent approaches.

\newpage

\subsection{Analysis of Heterogeneous Reasoning Strategies}~\label{app:HB_analysis}
\begin{figure}[!t]
    \centering
    \includegraphics[width=1.0\linewidth]{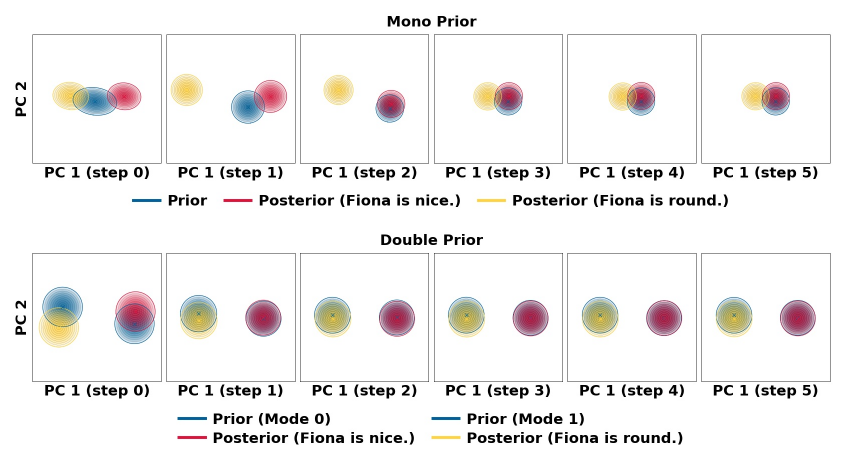}
    \caption{PCA visualization of prior and posterior latent distributions for a HiddenBranch example with two valid reasoning branches. Top: unimodal prior. Bottom: two-component strategy-conditioned prior.}
    \label{fig:appendix_pca_double}
\end{figure}

\begin{figure}[!t]
    \centering
    \includegraphics[width=1.0\linewidth]{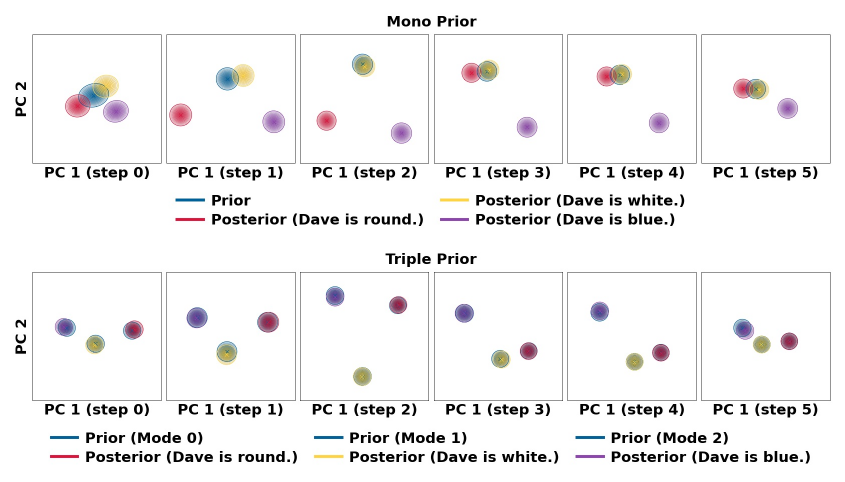}
    \caption{PCA visualization of prior and posterior latent distributions for a HiddenBranch example with three valid reasoning branches. Top: unimodal prior. Bottom: three-component strategy-conditioned prior.}
    \label{fig:appendix_pca_triple}
\end{figure}

\begin{figure}[!t]
    \centering
    \includegraphics[width=1.0\linewidth]{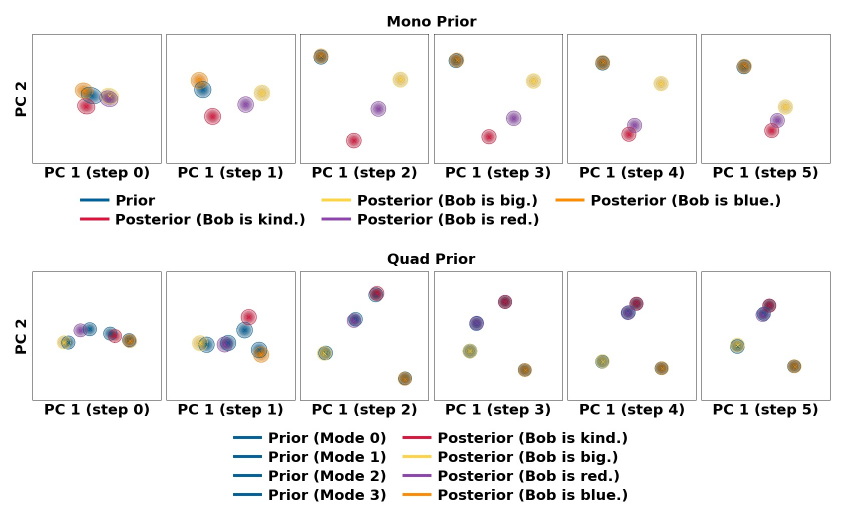}
    \caption{PCA visualization of prior and posterior latent distributions for a HiddenBranch example with four valid reasoning branches. Top: unimodal prior. Bottom: four-component strategy-conditioned prior.}
    \label{fig:appendix_pca_quad}
\end{figure}

\paragraph{Multimodal latent reasoning.} To further examine how PLRM represents heterogeneous reasoning trajectories, we visualize the learned prior and posterior distributions in the latent space. Figures~\ref{fig:appendix_pca_double}--\ref{fig:appendix_pca_quad} show representative HiddenBranch examples with two, three, and four valid
reasoning branches, respectively. For each example, we compare a unimodal prior with a strategy-conditioned prior whose number of components matches the number of valid reasoning branches. As shown in Figure~\ref{fig:appendix_pca_double}, when two distinct posterior modes are induced by different valid reasoning trajectories, a single Gaussian prior must account for both modes simultaneously. In contrast, the
strategy-conditioned prior maintains separate components corresponding to the different posterior modes across latent reasoning steps. Figures~\ref{fig:appendix_pca_triple} and \ref{fig:appendix_pca_quad} show that the same behavior extends to settings with three and four valid reasoning branches. As the number of heterogeneous trajectories increases, the posterior exhibits multiple distinct modes, while the corresponding strategy-conditioned prior preserves multiple components rather than representing all trajectories with a single distribution. These visualizations are consistent with the motivation in Section~3.3: the discrete strategy variable separates heterogeneous reasoning modes, allowing the continuous latent variables to model variation within each mode.

\begin{figure}[h]
    \centering
    \includegraphics[width=1.0\linewidth]{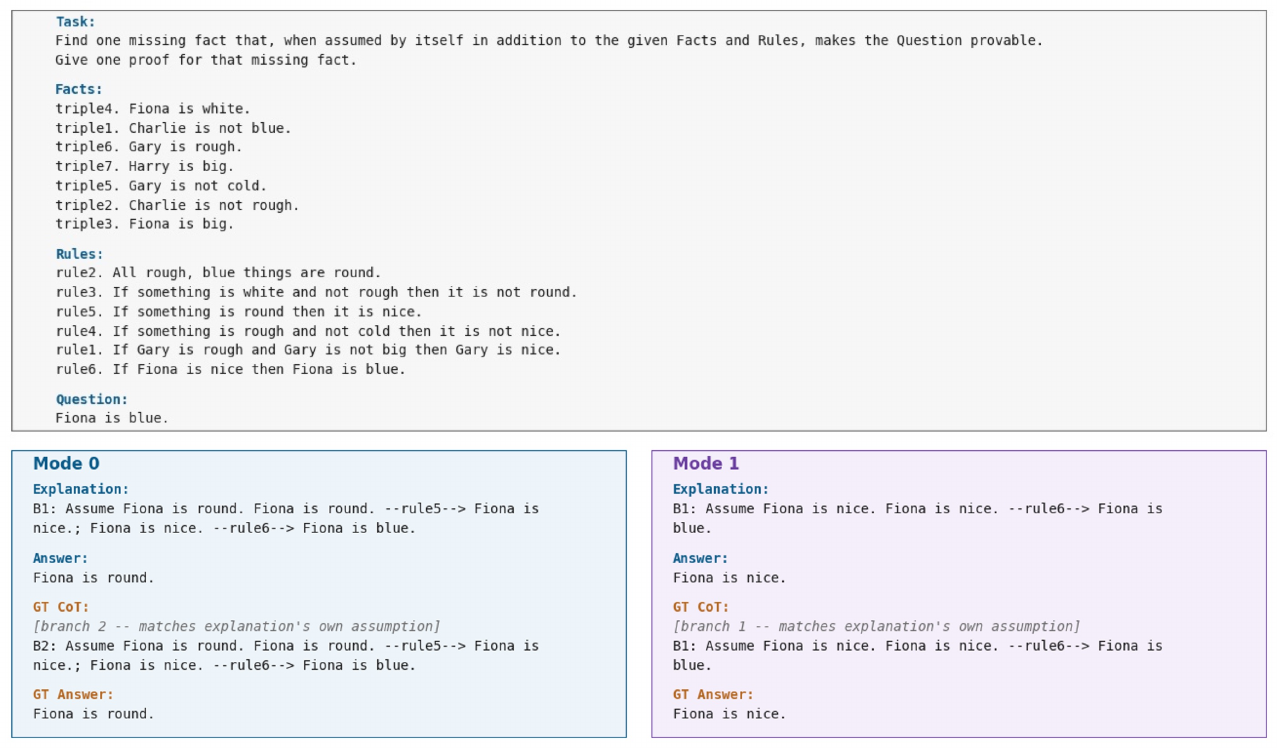}
    \caption{Qualitative HiddenBranch example with two valid reasoning branches generated from different strategy modes.}
    \label{fig:appendix_hiddenbranch_double}
\end{figure}

\begin{figure}[h]
    \centering
    \includegraphics[width=1.0\linewidth]{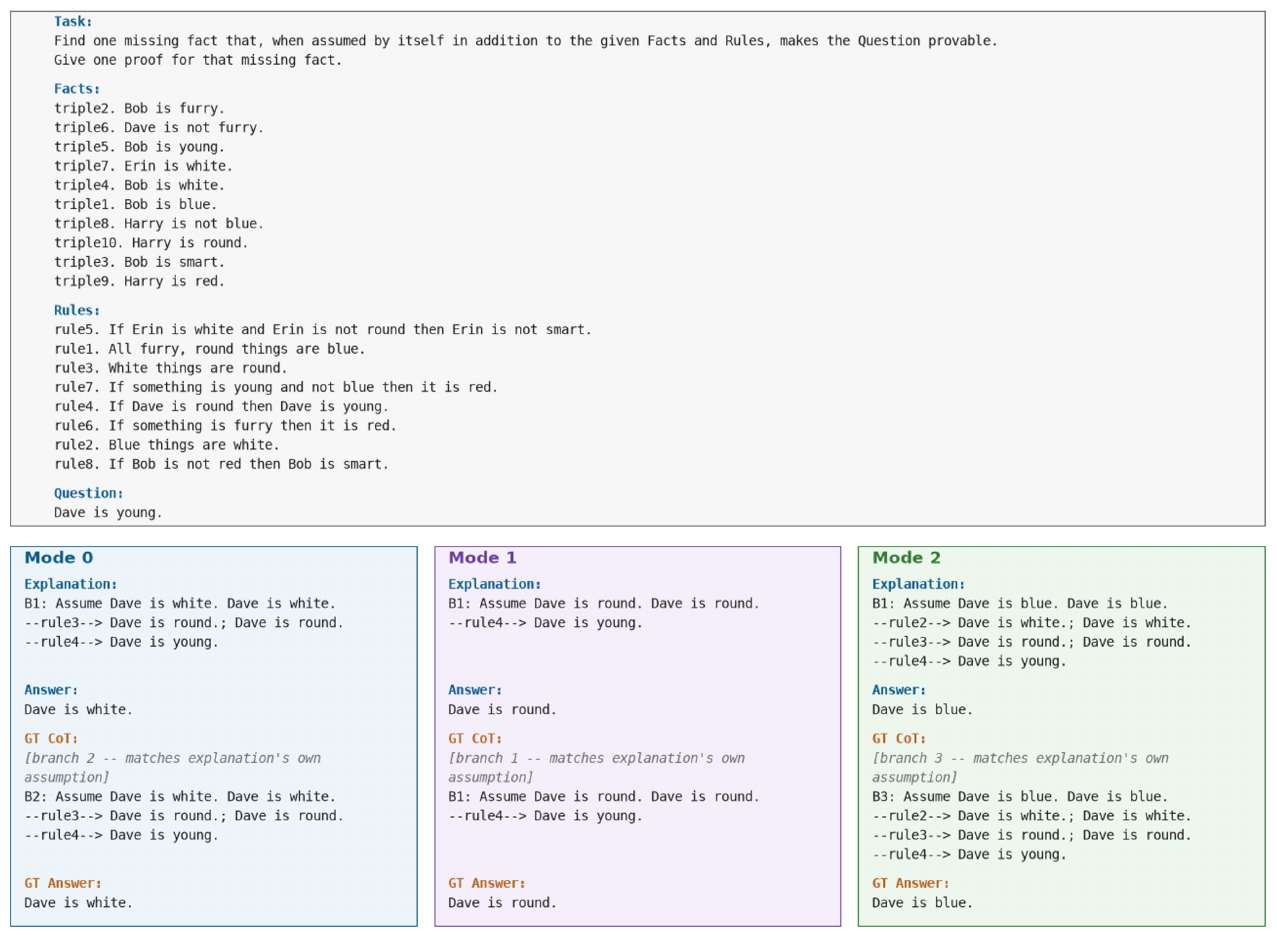}
    \caption{Qualitative HiddenBranch example with three valid reasoning branches generated from different strategy modes.}
    \label{fig:appendix_hiddenbranch_triple}
\end{figure}

\begin{figure}[h]
    \centering
    \includegraphics[width=1.0\linewidth]{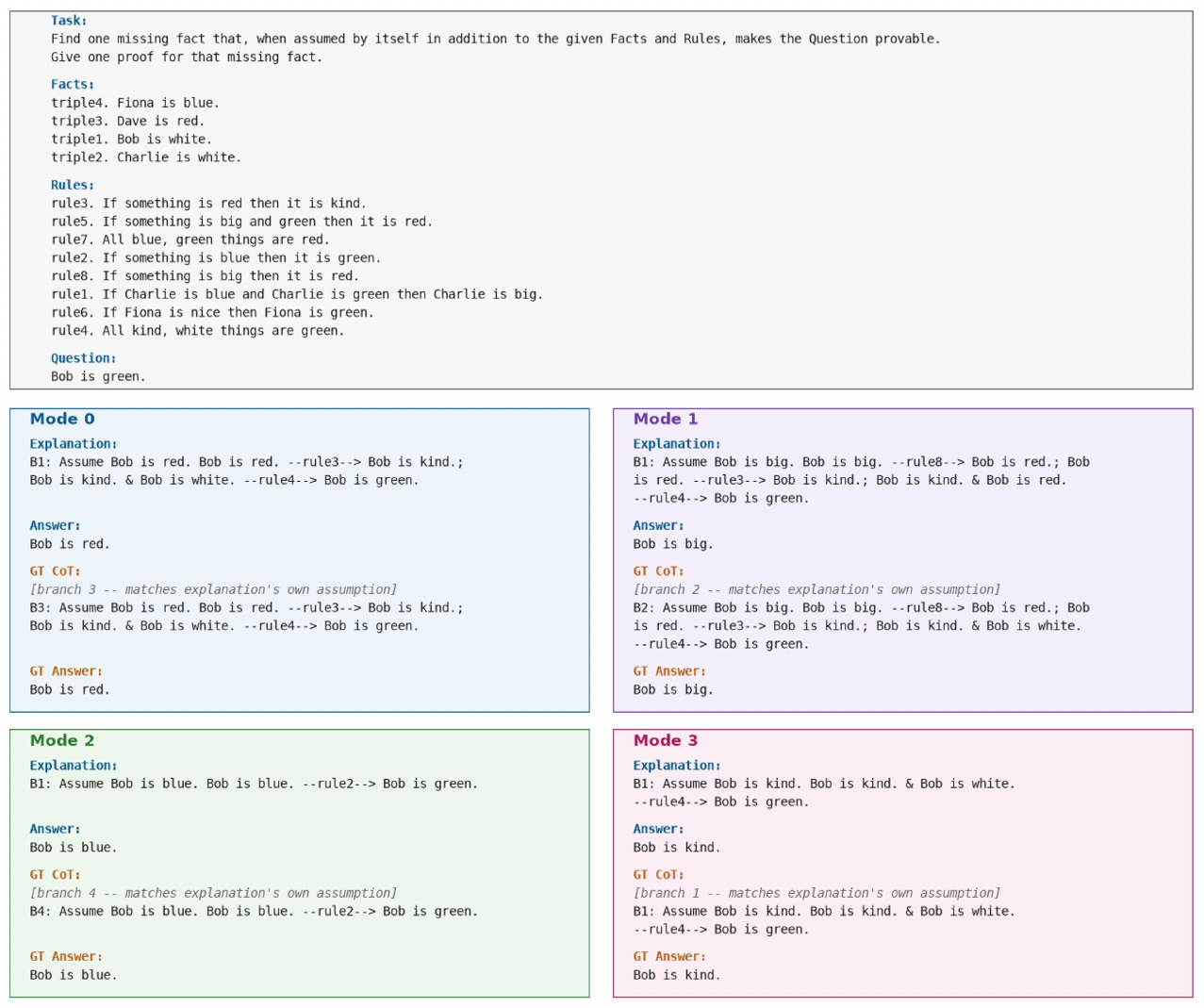}
    \caption{Qualitative HiddenBranch example with four valid reasoning branches generated from different strategy modes.}
    \label{fig:appendix_hiddenbranch_quad}
\end{figure}

\paragraph{Qualitative strategy-conditioned generations.}
To examine whether the separated latent modes correspond to meaningful differences in reasoning behavior, we further inspect generations conditioned on individual strategy modes. Figures~\ref{fig:appendix_hiddenbranch_double}--\ref{fig:appendix_hiddenbranch_quad} present representative HiddenBranch examples with two, three, and four valid reasoning branches, respectively. Given the same input, different strategy modes generate distinct valid missing facts together with their corresponding reasoning trajectories and answers. In the two-branch example (Figure~\ref{fig:appendix_hiddenbranch_double}), the two strategy modes recover different valid missing facts and follow different proof paths for the same query. Figures~\ref{fig:appendix_hiddenbranch_triple} and
\ref{fig:appendix_hiddenbranch_quad} extend this observation to three- and four-branch settings, where individual modes similarly generate distinct valid solutions. Importantly, the diversity across modes reflects differences in the underlying reasoning paths rather than merely variations in surface-level
verbalization. Together with the latent-space visualizations above, these examples suggest that the learned strategy variable captures semantically distinct reasoning modes, while the continuous latent trajectory represents the reasoning process associated with each mode.

\end{document}